\documentclass[11pt, twoside, letter]{article}
\usepackage[margin = 1in]{geometry}
\usepackage{array}
\usepackage{subcaption}
\usepackage{amsthm}
\usepackage{amsmath}
\usepackage{amsfonts}
\usepackage{microtype}
\usepackage{graphicx}
\usepackage{comment}
\usepackage{booktabs} 
\newtheorem{theorem}{Theorem}[section]
\newtheorem{lemma}[theorem]{Lemma}
\newtheorem{proposition}[theorem]{Proposition}

\theoremstyle{definition}
\newtheorem{definition}[theorem]{Definition}

\usepackage{color}
\usepackage{graphicx}
\usepackage{pgfplots}

\usepackage{hyperref}
\usepackage{xr}
\theoremstyle{remark}
\newtheorem{remark}[theorem]{Remark}

\newcommand{\abs}[1]{\ensuremath{\left|#1\right|}}

\renewcommand{\vec}[1]{\mathbf{#1}}

\newcommand{\argmin}{\mathop{\mathbf{argmin}}}
\newcommand{\argmax}{\mathop{\mathbf{argmax}}}
\usepackage{float}

\begin{document}

\title{Last iterate convergence in no-regret learning: constrained min-max optimization for convex-concave landscapes}

\author{Qi Lei\\UT Austin\\leiqi@ices.utexas.edu
\and Sai Ganesh Nagarajan\\SUTD\\sganesh.22@gmail.com
\and Ioannis Panageas\\SUTD\\ioannis@sutd.edu.sg
\and Xiao Wang\\SUTD\\xiao\_wang@sutd.edu.sg
}


\date{}
\maketitle

\begin{abstract}
In a recent series of papers it has been established that variants of Gradient Descent/Ascent and Mirror Descent exhibit last iterate convergence in convex-concave zero-sum games. Specifically, \cite{DISZ17, LiangS18} show last iterate convergence of the so called ``Optimistic Gradient Descent/Ascent" for the case of \textit{unconstrained} min-max optimization. Moreover, in \cite{Metal} the authors show that Mirror Descent with an extra gradient step displays last iterate convergence for convex-concave problems (both constrained and unconstrained), though their algorithm does not follow the online learning framework; it uses extra information rather than \textit{only} the history to compute the next iteration.
In this work, we show that "Optimistic Multiplicative-Weights Update (OMWU)" which follows the no-regret online learning framework, exhibits last iterate convergence locally for convex-concave games, generalizing the results of \cite{DP19} where last iterate convergence of OMWU was shown only for the \textit{bilinear case}. We complement our results with experiments that indicate fast convergence of the method.
\end{abstract}

\section{Introduction}
In classic (normal form) zero-sum games, one has to compute two probability vectors $\vec{x}^* \in \Delta_n, \vec{y}^* \in \Delta_m$ \footnote{$\Delta_n$ denotes the simplex of size $n$.} that consist an equilibrium of the following problem
\begin{align}
\min_{\vec{x}\in \Delta_n}\max_{\vec{y} \in \Delta_m} \vec{x}^{\top} A \vec{y}, \label{eq:min-max problem}
\end{align}
where $A$ is $n \times m$ real matrix (called payoff matrix). Here $\vec{x}^{\top}A\vec{y}$ represents the payment of the $\vec{x}$ player to the $\vec{y}$ player under choices of strategies by the two players and is a \textit{bilinear} function.

Arguably, one of the most celebrated theorems and a founding stone in Game Theory, is the minimax theorem by Von Neumann ~\cite{VN28}. It states that
\begin{align}
\min_{\vec{x}\in \Delta_n}\max_{\vec{y} \in \Delta_m} f(\vec{x},\vec{y}) = \max_{\vec{y} \in \Delta_m} \min_{\vec{x}\in \Delta_n} f(\vec{x},\vec{y}), \label{eq:minimax theorem}
\end{align}
where $f: \Delta_n \times \Delta_m \to \mathbb{R}$ is convex in $\vec{x}$, concave in $\vec{y}$. The aforementioned result holds for any convex compact sets $\mathcal{X} \subset \mathbb{R}^n$ and $\mathcal{Y} \subset \mathbb{R}^m$. The min-max theorem reassures us that an equilibrium always exists in the bilinear game (\ref{eq:min-max problem}) or its convex-concave analogue (again $f(\vec{x},\vec{y})$ is interpreted as the payment of the $\vec{x}$ player to the $\vec{y}$ player). An equilibrium is a pair of randomized strategies $(\vec{x}^*,\vec{y}^*)$ such that neither player can improve their payoff by unilaterally changing their distribution.

Soon after the appearance of the minimax theorem, research was focused on dynamics for solving min-max optimization problems by having the min and max players of~\eqref{eq:min-max problem} run a simple online learning procedure. In the online learning framework, at time $t$, each player chooses a probability distribution ($\vec{x}^t, \vec{y}^t$ respectively) simultaneously depending \textit{only} on the past choices of both players (i.e., $\vec{x}^{1},...,\vec{x}^{t-1},\vec{y}^1,...,\vec{y}^{t-1}$) and experiences payoff that depends on choices $\vec{x}^t,\vec{y}^t$.

An early method, proposed by Brown~\cite{B51} and analyzed by Robinson~\cite{R51}, was fictitious play. Later on, researchers discover several learning robust algorithms converging to minimax equilibrium at faster rates, see \cite{C06}. This class of learning algorithms, are the so-called ``no-regret'' and include Multiplicative Weights Update method \cite{arora12} and Follow the regularized leader.

\subsection{Average Iterate Convergence}
Despite the rich literature on no-regret learning, most of the known results have the feature that min-max equilibrium is shown to be attained only by the time \textit{average}.  This means that the trajectory of a no-regret learning method $(\vec{x}^t,\vec{y}^t)$ has the property that  ${1 \over t}\sum_{\tau \le t}{\vec{x}^{\tau}}^{\top} A \vec{y}^{\tau}$ converges to the equilibrium of~\eqref{eq:min-max problem}, as $t \rightarrow \infty$. Unfortunately that does not mean that the last iterate $(\vec{x}^t,\vec{y}^t)$ converges to an equilibrium, it commonly diverges or cycles. One such example is the well-known Multiplicative Weights Update Algorithm, the time average of which is known to converge to an equilibrium, but the actual trajectory cycles towards the boundary of the simplex (\cite{BP18}). This is even true for the vanilla Gradient Descent/Ascent, where one can show for even bilinear landscapes (unconstrained case) last iterate fails to converge \cite{DISZ17}.

Motivated by the training of Generative Adversarial Networks (GANs), the last couple of years researchers have focused on designing and analyzing procedures that exhibit \textit{last iterate} convergence (or \text{pointwise} convergence) for zero-sum games. This is crucial for training GANs, the landscapes of which are typically non-convex non-concave and averaging now as before does not give much guarantees (e.g., note that Jensen's inequality is not applicable anymore). In \cite{DISZ17, LiangS18} the authors show that a variant of Gradient Descent/Ascent, called Optimistic Gradient Descent/Ascent has last iterate convergence for the case of bilinear functions $\vec{x}^{\top}A\vec{y}$ where $\vec{x} \in \mathbb{R}^n$ and $\vec{y} \in \mathbb{R}^m$ (this is called the unconstrained case, since there are no restrictions on the vectors). Later on, \cite{DP19} generalized the above result with simplex constraints, where the online method that the authors analyzed was Optimistic Multiplicative Weights Update. In \cite{Metal}, it is shown that Mirror Descent with extra gradient computation converges pointwise for a class of zero-sum games that includes the convex-concave setting (with arbitrary constraints), though their algorithm does not fit in the online no-regret framework since it uses information twice about the payoffs before it iterates. Last but not least there have appeared other works that show pointwise convergence for other settings (see \cite{PPP17, DP18} and \cite{jake19} and references therein) to stationary points (but not local equilibrium solutions).

\subsection{Main Results}
In this paper, we focus on the min-max optimization problem
\begin{align}
\min_{\vec{x}\in \Delta_n}\max_{\vec{y} \in \Delta_m} f(\vec{x}, \vec{y}), \label{eq:min-max problemgeneral}
\end{align}
where $f$ is a convex-concave function (convex in $\vec{x}$, concave in $\vec{y}$). We analyze the no-regret online algorithm Optimistic Multiplicative Weights Update (OMWU). OMWU is an instantiation of the Optimistic Follow the Regularized Leader (OFTRL) method with entropy as a regularizer (for both players, see Preliminaries section for the definition of OMWU). 

We prove that OMWU exhibits local last iterate convergence, generalizing the result of \cite{DP19} and proving an open question of \cite{SALS15} (for convex-concave games).
Formally, our main theorem is stated below:
\begin{theorem}[Last iterate convergence of OMWU]\label{thm:main} Let $f: \Delta_n \times \Delta_m \to \mathbb{R}$ be a twice differentiable function $f(\vec{x},\vec{y})$ that is convex in $\vec{x}$ and concave in $\vec{y}$. Assume that there exists an equilibrium $(\vec{x}^*,\vec{y}^*)$ that satisfies Remark \ref{rem:support}. It holds that for sufficiently small stepsize, there exists a neighborhood $U \subseteq \Delta_n \times \Delta_m$ of $(\vec{x}^*,\vec{y}^*)$ such that for all for all initial conditions $(\vec{x}^0,\vec{y}^0), (\vec{x}^1,\vec{y}^1) \in U$, OMWU exhibits last iterate (pointwise) convergence, i.e.,
\[\lim_{t\to \infty} (\vec{x}^t,\vec{y}^t) = (\vec{x}^*,\vec{y}^*), \] where $(\vec{x}^t,\vec{y}^t)$ denotes the $t$-th iterate of OMWU.
\end{theorem}

Moreover, we complement our theoretical findings with experimental analysis of the procedure. The experiments on KL-divergence indicate that the results should hold globally.


\subsection{Structure and Technical Overview}

We present the structure of the paper and a brief technical overview.

\textbf{Section 2} provides necessary definitions, the explicit form of OMWU derived from OFTRL with entropy regularizer, and some existing results on dynamical systems.

\textbf{Section 3} is the main technical part, i.e, the computation and spectral analysis of the Jacobian matrix of OMWU dynamics.
The stability analysis, the understanding of the local behavior and the local convergence guarantees of OMWU rely on the spectral analysis of the computed Jacobian matrix. The techniques for bilinear games (as in \cite{DP19}) are no longer valid in convex-concave games. Allow us to explain the differences from \cite{DP19}. In general, one cannot expect a trivial generalization from linear to non-linear scenarios. The properties of bilinear games are fundamentally different from that of convex-concave games, and this makes the analysis much more challenging in the latter. The key result of spectral analysis in \cite{DP19} is in a lemma (Lemma B.6) which states that a skew symmetric\footnote{$A$ is skew symmetric if $A^{\top} = -A$.} has imaginary eigenvalues. Skew symmetric matrices appear since in bilinear cases there are terms that are linear in $\vec{x}$ and linear in $\vec{y}$ but no higher order terms in $\vec{x}$ or $\vec{y}$. However, the skew symmetry has no place in the case of convex-concave landscapes and the Jacobian matrix of OMWU is far more complicated. One key technique to overcome the lack of skew symmetry is the use of Ky Fan inequality \cite{kyfan} which states that the sequence of the eigenvalues of $\frac{1}{2}(W+W^{\top})$ majorizes the real part of the sequence of the eigenvalues of W for any square matrix W (see Lemma 3.1).

\textbf{Section 4} focuses on numerical experiments to understand how the problem size and the choice of learning rate affect the performance of our algorithm. We observe that our algorithm is able to achieve global convergence invariant to the choice of learning rate, random initialization or problem size. As comparison, the latest popularized (projected) optimistic gradient descent ascent is much more sensitivity to the choice of hyperparameter.
Due to space constraint, the detailed calculation of the Jacobian matrix (general form and at fixed point) of OMWU are left in Appendix.

\paragraph{Notation}
The boldface $\vec{x}$ and $\vec{y}$ denote the vectors in $\Delta_n$ and $\Delta_m$. $\vec{x}^t$ denotes the $t$-th iterate of the dynamical system. The letter $J$ denote the Jacobian matrix. $\vec{I}$, $\vec{0}$ and $\vec{1}$ are preserved for the identity, zero matrix and the vector with all the entries equal to 1. The support of $\vec{x}$ is the set of indices of $x_i$ such that $x_i\ne 0$, denoted by $\textrm{Supp}(\vec{x})$. $(\vec{x}^*,\vec{y}^*)$ denotes the optimal solution for minimax problem. $[n]$ denote the set of integers $\{1,...,n\}$.
\section{Preliminaries}
In this section, we present some background that will be used later.
\subsection{Equilibria for Constrained Minimax}
From Von Neumann's minimax theorem, one can conclude that the problem $\min_{\vec{x} \in \Delta_n} \max_{\vec{y} \in \Delta} f(\vec{x},\vec{y})$ has always an equilibrium $(\vec{x}^*,\vec{y}^*)$ with $f(\vec{x}^*,\vec{y}^*)$ be unique. Moreover from KKT conditions (as long as $f$ is twice differentiable), such an equilibrium must satisfy the following ($\vec{x}^*$ is a local minimum for fixed $\vec{y} = \vec{y}^*$ and $\vec{y}^*$ is a local maximum for fixed $\vec{x} = \vec{x}^*$):

\begin{definition}[KKT conditions] Formally, it holds
\begin{equation}\label{eq:KKT}
\begin{array}{l}
\vec{x}^* \in \Delta_n\\
x_{i}^*>0 \Rightarrow \frac{\partial f}{\partial x_{i}}(\vec{x}^*,\vec{y}^*) = \sum_{j=1}^n x_{j}^*\frac{\partial f}{\partial x_{j}}(\vec{x}^*,\vec{y}^*)\\
x_{i}^*=0 \Rightarrow \frac{\partial f}{\partial x_{i}}(\vec{x}^*,\vec{y}^*) \geq \sum_{j=1}^n x_{j}^*\frac{\partial f}{\partial x_{j}}(\vec{x}^*,\vec{y}^*)\\ \textrm{ for player }\vec{x},\\
\vec{y}^* \in \Delta_m\\
y_{i}^*>0 \Rightarrow \frac{\partial f}{\partial y_{i}}(\vec{x}^*,\vec{y}^*) = \sum_{j=1}^m y_{j}^*\frac{\partial f}{\partial y_{j}}(\vec{x}^*,\vec{y}^*)\\
y_{i}^*=0 \Rightarrow \frac{\partial f}{\partial y_{i}}(\vec{x}^*,\vec{y}^*) \leq \sum_{j=1}^m y_{j}^*\frac{\partial f}{\partial y_{j}}(\vec{x}^*,\vec{y}^*)\\ \textrm{ for player }\vec{y}.
\end{array}
\end{equation}
\begin{remark}[No degeneracies]\label{rem:support}
For the rest of the paper we assume no degeneracies, i.e., the KKT inequalities hold strictly (in the case a strategy is played with zero probability for each player). Moreover, it is easy to see that since $f$ is convex concave and twice differentiable, then $\nabla^2_{\vec{x}\vec{x}} f$ (part of the Hessian that involves $\vec{x}$ variables) is positive semi-definite and $\nabla^2_{\vec{y}\vec{y}} f$ (part of the Hessian that involves $\vec{y}$ variables) is negative semi-definite. Furthermore we assume that the submatrix of the Hessian of $f$ computed at the minmax point $(\vec{x}^*,\vec{y}^*)$ and is generated by removing the rows and columns outside the support of $\vec{x}^*,\vec{y}^*$ is invertible.    
\end{remark}
\end{definition}

\subsection{Optimistic Multiplicative Weights Update}\label{sec:omwu}
The equations of Optimistic Follow-the-Regularized-Leader (OFTRL) applied to a problem \\$\min_{\vec{x} \in \mathcal{X}} \max_{\vec{y} \in \mathcal{Y}} f(\vec{x},\vec{y})$ with regularizers (strongly convex functions) $h_1(\vec{x}), h_2 (\vec{y})$ (for player $\vec{x}, \vec{y}$ respectively) and $\mathcal{X} \subset \mathbb{R}^n, \mathcal{Y} \subset \mathbb{R}^m$ is given below (see \cite{DISZ17}):
\begin{align*}
\vec{x}^{t+1}&=\argmin_{\vec{x}\in \mathcal{X}}\{\eta\sum_{s=1}^t \vec{x}^{\top}\nabla_{\vec{x}}f(\vec{x}^s,\vec{y}^s)+\underbrace{\eta \vec{x}^{\top}\nabla_{\vec{x}} f(\vec{x}^t,\vec{y}^t)}_{\textrm{optimistic term}}+h_1(\vec{x})\}
\\
\vec{y}^{t+1}&=\argmax_{\vec{y}\in\mathcal{Y}}\{\eta\sum_{s=1}^t \vec{y}^{\top} \nabla_{\vec{y}} f(\vec{x}^s,\vec{y}^s)+\underbrace{\eta \vec{y}^{\top}\nabla_{\vec{y}} f(\vec{x}^t,\vec{y}^t)}_{\textrm{optimistic term}}-h_2(\vec{y})\}.
\end{align*}
$\eta$ is called the \textit{stepsize} of the online algorithm. OFTRL is uniquely defined if $f$ is convex-concave and domains $\mathcal{X}$ and $\mathcal{Y}$ are convex. For simplex constraints and entropy regularizers, i.e., $h_1(\vec{x})=\sum_ix_i\ln x_i, h_2(\vec{y})=\sum_iy_i\ln y_i$, we can solve for the explicit form of OFTRL using KKT conditions, the update rule is the Optimistic Multiplicative Weights Update (OMWU) and is described as follows:
\begin{align*}
x_i^{t+1}&=x_i^t\frac{e^{-2\eta\frac{\partial f}{\partial x_i}(\vec{x}^t,\vec{y}^t)+\eta\frac{\partial f}{\partial x_i}(\vec{x}^{t-1},\vec{y}^{t-1})}}{\sum_k x_k^te^{-2\eta\frac{\partial f}{\partial x_k}(\vec{x}^t,\vec{y}^t)+\eta\frac{\partial f}{\partial x_k}(\vec{x}^{t-1},\vec{y}^{t-1})}}
\\
&\text{for all } i\in[n],
\\
y_i^{t+1}&=y_i^t\frac{e^{2\eta\frac{\partial f}{\partial y_i}(\vec{x}^t,\vec{y}^t)-\eta\frac{\partial f}{\partial y_i}(\vec{x}^{t-1},\vec{y}^{t-1})}}{\sum_ky_k^te^{2\eta\frac{\partial f}{\partial y_j}(\vec{x}^t,\vec{y}^t)-\eta\frac{\partial f}{\partial y_k}(\vec{x}^{t-1},\vec{y}^{t-1})}}
\\
&\text{for all } i\in[m].
\end{align*}

\subsection{Fundamentals of Dynamical Systems}
We conclude Preliminaries section with some basic facts from dynamical systems.
\begin{definition}
A recurrence relation of the form $\vec{x}^{t+1}=w(\vec{x}^t)$ is a discrete time dynamical system, with update rule $w:\mathcal{S}\rightarrow\mathcal{S}$ where $\mathcal{S}$ is a subset of $\mathbb{R}^{k}$ for some positive integer $k$. The point $\vec{z}\in\mathcal{S}$ is called a \emph{fixed point} if $w(\vec{z})=\vec{z}$.
\end{definition}
\begin{remark}\label{rem:fixedpoint} Using KKT conditions (\ref{eq:KKT}), it is not hard to observe that an equilibrium point $(\vec{x}^*,\vec{y}^*)$ must be a fixed point of the OMWU algorithm, i.e., if $(\vec{x}^t,\vec{y}^t) = (\vec{x}^{t-1},\vec{y}^{t-1}) = (\vec{x}^*,\vec{y}^*)$ then $(\vec{x}^{t+1},\vec{y}^{t+1}) = (\vec{x}^*,\vec{y}^*)$.
\end{remark}
\begin{proposition}[\cite{G07}]\label{prop:contract}
Assume that $w$ is a differentiable function and the Jacobian of the update rule $w$ at a fixed point $\vec{z}^*$ has spectral radius less than one. It holds that there exists a neighborhood $U$ around $\vec{z}^*$ such that for all $\vec{z}^0\in U$, the dynamics $\vec{z}^{t+1}=w(\vec{z}^t)$ converges to $\vec{z}^*$, i.e. $\lim_{n\rightarrow\infty}w^n(\vec{z}^0)=\vec{z}^*$ \footnote{$w^n$ denotes the composition of $w$ with itself $n$ times.}. $w$ is called a contraction mapping in $U$.
\end{proposition}
Note that we will make use of Proposition \ref{prop:contract} to prove our Theorem \ref{thm:main} (by proving that the Jacobian of the update rule of OMWU has spectral radius less than one).

\section{Last iterate convergence of OMWU}
In this section, we prove that OMWU converges pointwise (exhibits last iterate convergence) if the initializations $(\vec{x}^0,\vec{y}^0),(\vec{x}^1,\vec{y}^1)$ belong in a neighborhood $U$ of the equilibrium $(\vec{x}^*,\vec{y}^*)$. 
\subsection{Dynamical System of OMWU}
We first express OMWU algorithm as a dynamical system so that we can use Proposition \ref{prop:contract}. The idea (similar to \cite{DP19}) is to lift the space to consist of four components $(\vec{x},\vec{y},\vec{z},\vec{w}$, in such a way we can include the history (current and previous step, see Section \ref{sec:omwu} for the equations).
First, we provide the update rule $g: \Delta_n \times \Delta_m \times \Delta_n \times \Delta_m \to \Delta_n \times \Delta_m \times \Delta_n \times \Delta_m$ of the lifted dynamical system and is given by
\[
g(\vec{x},\vec{y},\vec{z},\vec{w})=(g_1,g_2,g_3,g_4)
\]
where $g_i=g_i(\vec{x},\vec{y},\vec{z},\vec{w})$ for $i\in[4]$ are defined as follows:
\begin{align}
g_{1,i}(\vec{x},\vec{y},\vec{z},\vec{w})&=x_i\frac{e^{-2\eta\frac{\partial f}{\partial x_i}(\vec{x},\vec{y})+\eta\frac{\partial f}{\partial z_i}(\vec{z},\vec{w})}}{\sum_kx_ke^{-2\eta\frac{\partial f}{\partial x_k}(\vec{x},\vec{y})+\eta\frac{\partial f}{\partial z_k}(\vec{z},\vec{w})}}, i\in[n]
\\
g_{2,i}(\vec{x},\vec{y},\vec{z},\vec{w})&=y_i\frac{e^{2\eta\frac{\partial f}{\partial y_i}(\vec{x},\vec{y})-\eta\frac{\partial f}{\partial w_i}(\vec{z},\vec{w})}}{\sum_ky_ke^{2\eta\frac{\partial f}{\partial y_k}(\vec{x},\vec{y})-\eta\frac{\partial f}{\partial w_k}(\vec{z},\vec{w})}},  i\in[m]
\\
g_3(\vec{x},\vec{y},\vec{z},\vec{w})&=\vec{x}\ \ \ \text{or} \ \ g_{3,i}(\vec{x},\vec{y},\vec{z},\vec{w})=x_i, \ i\in[n]
\\
g_4(\vec{x},\vec{y},\vec{z},\vec{w})&=\vec{y}\ \ \ \text{or} \ \ g_{4,i}(\vec{x},\vec{y},\vec{z},\vec{w})=y_i,\ i\in[m].
\end{align}
Then the dynamical system of OMWU can be written in compact form as
\[
(\vec{x}_{t+1},\vec{y}_{t+1},\vec{x}_t,\vec{y}_t)=g(\vec{x}_t,\vec{y}_t,\vec{x}_{t-1},\vec{y}_{t-1}).
\]
In what follows, we will perform spectral analysis on the Jacobian of the function $g$, computed at the fixed point $(\vec{x}^*,\vec{y}^*)$. Since $g$ has been lifted, the fixed point we analyze is $(\vec{x}^*,\vec{y}^*,\vec{x}^*,\vec{y}^*)$ (see Remark \ref{rem:fixedpoint}).
By showing that the spectral radius is less than one, our Theorem \ref{thm:main} follows by Proposition \ref{prop:contract}. The computations of the Jacobian of $g$ are deferred to the supplementary material.

\subsection{Spectral Analysis}
Let $(\vec{x}^*,\vec{y}^*)$ be the equilibrium of min-max problem (\ref{eq:minimax theorem}). Assume $i \notin \textrm{Supp}(\vec{x}^*)$, i.e., $x_i^*=0$ then (see equations at the supplementary material, section A) \[\frac{\partial g_{1,i}}{\partial x_i}(\vec{x}^*,\vec{y}^*,\vec{x}^*,\vec{y}^*) = \frac{e^{-\eta \frac{\partial f}{\partial x_i}(\vec{x}^*,\vec{y}^*)} }{\sum_{t=1}^n x^*_t e^{-\eta \frac{\partial f}{\partial x_t}(\vec{x}^*,\vec{y}^*)} } \] and all other partial derivatives of $g_{1,i}$ are zero, thus $\frac{e^{-\eta \frac{\partial f}{\partial x_i}(\vec{x}^*,\vec{y}^*)} }{\sum_{t=1}^n x^*_t e^{-\eta \frac{\partial f}{\partial x_t}(\vec{x}^*,\vec{y}^*)} }$ is an eigenvalue of the Jacobian computed at $(\vec{x}^*,\vec{y}^*,\vec{x}^*,\vec{y}^*)$. This is true because the row of the Jacobian that corresponds to $g_{1,i}$ has zeros everywhere but the diagonal entry. Moreover because of the degeneracy assumption of KKT conditions (see Remark \ref{rem:support}), it holds that 
\[\frac{e^{-\eta \frac{\partial f}{\partial x_i}(\vec{x}^*,\vec{y}^*)} }{\sum_{t=1}^n x^*_t e^{-\eta \frac{\partial f}{\partial x_t}(\vec{x}^*,\vec{y}^*)} }<1.\] Similarly, it holds for $j \notin \textrm{Supp}(\vec{y}^*)$ that 
\[\frac{\partial g_{2,j}}{\partial y_j}(\vec{x}^*,\vec{y}^*,\vec{x}^*,\vec{y}^*) = \frac{e^{\eta \frac{\partial f}{\partial y_j}(\vec{x}^*,\vec{y}^*)} }{\sum_{t=1}^m y^*_t e^{\eta \frac{\partial f}{\partial y_t}(\vec{x}^*,\vec{y}^*)} }<1\]
 (again by Remark \ref{rem:support}) and all other partial derivatives of $g_{2,j}$ are zero, therefore $\frac{e^{\eta \frac{\partial f}{\partial y_j}(\vec{x}^*,\vec{y}^*)} }{\sum_{t=1}^m y^*_t e^{\eta \frac{\partial f}{\partial y_t}(\vec{x}^*,\vec{y}^*)} }$ is an eigenvalue of the Jacobian computed at $(\vec{x}^*,\vec{y}^*,\vec{x}^*,\vec{y}^*)$.

We focus on the submatrix of the Jacobian of $g$ computed at $(\vec{x}^*,\vec{y}^*,\vec{x}^*,\vec{y}^*)$ that corresponds to the non-zero probabilities of $\vec{x}^*$ and $\vec{y}^*$. We denote $D_{\vec{x}^*}$ to be the diagonal matrix of size $\abs{\text{Supp}(\vec{x}^*)}\times\abs{\text{Supp}(\vec{x}^*)}$ that has on the diagonal the nonzero entries of $\vec{x}^*$ and similarly we define $D_{\vec{y}^*}$ of size $\abs{\text{Supp}(\vec{y}^*)}\times\abs{\text{Supp}(\vec{y}^*)}$. For convenience, let us denote $k_x:=\abs{\text{Supp}(\vec{x}^*)}$ and $k_y:=\abs{\text{Supp}(\vec{y}^*)}$. The Jacobian submatrix is the following
\[
J=
\left[
\begin{array}{cccc}
A_{11} & A_{12} & A_{13} & A_{14}
\\
A_{21} & A_{22} & A_{23} & A_{24}
\\
\vec{I}_{k_x \times k_x} & \vec{0}_{k_x \times k_y} & \vec{0}_{k_x \times k_x} & \vec{0}_{k_x \times k_y}
\\
\vec{0}_{k_y \times k_x} & \vec{I}_{k_y \times k_y} & \vec{0}_{k_y \times k_x} & \vec{0}_{k_y \times k_y}
\end{array}
\right]
\]
where
\begin{align}
\begin{split}
A_{11}&=\vec{I}_{k_x \times k_x}-D_{\vec{x}^*}\vec{1}_{k_x}\vec{1}_{k_x}^{\top}-2\eta D_{\vec{x}^*}(\vec{I}_{k_x \times k_x}-\vec{1}_{k_x}\vec{x}^{*\top})\nabla_{\vec{x}\vec{x}}^2f\\
A_{12}&=-2\eta D_{\vec{x}^*}(\vec{I}_{k_x \times k_x}-\vec{1}_{k_x}\vec{x}^{*\top})\nabla_{\vec{x}\vec{y}}^2f\\
A_{13}&=\eta D_{\vec{x}^*}(\vec{I}_{k_x \times k_x}-\vec{1}_{k_x}\vec{x}^{*\top})\nabla_{\vec{x}\vec{x}}^2f\\
A_{14}&=\eta D_{\vec{x}^{*}}(\vec{I}_{k_x \times k_x}-\vec{1}_{k_x}\vec{x}^{*\top})\nabla_{\vec{x}\vec{y}}^2f\\
A_{21}&=2\eta D_{\vec{y}^{*}}(\vec{I}_{k_y \times k_y}-\vec{1}_{k_y}\vec{y}^{*\top})\nabla_{\vec{y}\vec{x}}^2f\\
A_{22}&=\vec{I}_{k_y \times k_y}-D_{\vec{y}^*}\vec{1}_{k_y}\vec{1}_{k_y}^{\top}+2\eta D_{\vec{y}^*}(\vec{I}_{k_y \times k_y}-\vec{1}_{k_y}\vec{y}^{*\top})\nabla_{\vec{y}\vec{y}}^2f\\
A_{23}&=-\eta D_{\vec{y}^*}(\vec{I}_{k_y \times k_y}-\vec{1}_{k_y}\vec{y}^{*\top})\nabla_{\vec{y}\vec{x}}^2f\\
A_{24}&=-\eta D_{\vec{y}^*}(\vec{I}_{k_y \times k_y}-\vec{1}_{k_y}\vec{y}^{*\top})\nabla_{\vec{y}\vec{y}}^2f.\\
\end{split}
\end{align}
We note that $\vec{I}, \vec{0}$ capture the identity matrix and the all zeros matrix respectively (the appropriate size is indicated as a subscript).  
The vectors $(\vec{1}_{k_x},\vec{0}_{k_y},\vec{0}_{k_x},\vec{0}_{k_y})$ and $(\vec{0}_{k_x},\vec{1}_{k_y},\vec{0}_{k_x},\vec{0}_{k_y})$ are left eigenvectors with eigenvalue zero for the above matrix. Hence, any right eigenvector $(\vec{v_x},\vec{v_y},\vec{v_z},\vec{v_w})$ should satisfy the conditions $\vec{1}^{\top}\vec{v_x}=0$ and $\vec{1}^{\top}\vec{v_y}=0$. Thus, every non-zero eigenvalue of the above matrix is also a non-zero eigenvalue of the matrix below:
\[
J_{\text{new}}=
\left[
\begin{array}{cccc}
B_{11} & A_{12} & A_{13} & A_{14}
\\
A_{21} & B_{22} & A_{23} & A_{24}
\\
\vec{I}_{k_x \times k_x} & \vec{0}_{k_x \times k_y} & \vec{0}_{k_x \times k_x} & \vec{0}_{k_x \times k_y}
\\
\vec{0}_{k_y \times k_x} & \vec{I}_{k_y \times k_y} & \vec{0}_{k_y \times k_x} & \vec{0}_{k_y \times k_y}
\end{array}
\right]
\]
where
\[
B_{11}=\vec{I}_{k_x \times k_x}-2\eta D_{\vec{x}^*}(\vec{I}_{k_x \times k_x}-\vec{1}_{k_x}\vec{x}^{*\top})\nabla_{\vec{x}\vec{x}}^2f,
\]
\[
B_{22}=\vec{I}_{k_y \times k_y}+2\eta D_{\vec{y}^*}(\vec{I}_{k_y \times k_y}-\vec{1}_{k_y}\vec{y}^{*\top})\nabla_{\vec{y}\vec{y}}^2f.
\]
The characteristic polynomial of $J_{\text{new}}$ is obtained by finding $\det(J_{\text{new}}-\lambda\vec{I})$. One can perform row/column operations on $J_{\text{new}}$ to calculate this determinant,  which gives us the following relation:
\[
\det(J_{\text{new}}-\lambda\vec{I}_{2k_x \times 2k_y})=\left(1-2\lambda\right)^{(k_x+k_y)}q\left(\frac{\lambda(\lambda-1)}{2\lambda-1}\right)
\]
where $q(\lambda)$ is the characteristic polynomial of the following matrix
\[
J_{\text{small}}=
\left[
\begin{array}{cccc}
B_{11}-\vec{I}_{k_x \times k_x} & A_{12}
\\
 A_{21} &  B_{22}-\vec{I}_{k_y \times k_y},
\end{array}
\right]
\]
and $B_{11},B_{12},A_{12},A_{21}$ are the aforementioned sub-matrices.
Notice that $J_{\text{small}}$ can be written as
\[
J_{\text{small}}=
2\eta
\left[
\begin{array}{cc}
-(D_{\vec{x}^*}-\vec{x}^{*}\vec{x}^{*\top}) & \vec{0}_{k_x \times k_y}
\\
\vec{0}_{k_y \times k_x} & (D_{\vec{y}^*}-\vec{y}^{*}\vec{y}^{*\top})
\end{array}
\right]H
\]
where,
\[
H=
\left[
\begin{array}{cc}
\nabla_{\vec{x}\vec{x}}^2f & \nabla_{\vec{x}\vec{y}}^2f
\\
\nabla_{\vec{y}\vec{x}}^2f & \nabla_{\vec{y}\vec{y}}^2f
\end{array}
\right]
\]
Notice here that $H$ is the Hessian matrix evaluated at the fixed point $(\vec{x}^*,\vec{y}^*)$, and is the appropriate sub-matrix restricted to the support of $\abs{\text{Supp}(\vec{y}^*)}$ and $\abs{\text{Supp}(\vec{x}^*)}$. Although, the Hessian matrix is symmetric, we would like to work with the following representation of $J_{\text{small}}$:
\[
J_{\text{small}}=
2\eta
\left[
\begin{array}{cc}
(D_{\vec{x}^*}-\vec{x}^{*}\vec{x}^{*\top}) & \vec{0}_{k_x \times k_y}
\\
\vec{0}_{k_y \times k_x} & (D_{\vec{y}^*}-\vec{y}^{*}\vec{y}^{*\top})
\end{array}
\right]H^{-}
\]
where,
\[
H^{-}=
\left[
\begin{array}{cc}
-\nabla_{\vec{x}\vec{x}}^2f & -\nabla_{\vec{x}\vec{y}}^2f
\\
\nabla_{\vec{y}\vec{x}}^2f & \nabla_{\vec{y}\vec{y}}^2f
\end{array}
\right]
\]

Let us denote any non-zero eigenvalue of $J_{\text{small}}$ by $\epsilon$ which may be a complex number. Thus $\epsilon$ is where $q(\cdot)$ vanishes and hence the eigenvalue of $J_{\text{new}}$ must satisfy the relation
\[
\frac{\lambda(\lambda-1)}{2\lambda-1}=\epsilon
\]

We are to now show that the magnitude of any eigenvalue of $J_{\text{new}}$ is strictly less than 1, i.e, $\abs{\lambda} < 1$. Trivially, $\lambda = \frac{1}{2}$ satisfies the above condition. Thus we need to show that the magnitude of $\lambda$ where $q(\cdot)$ vanishes is strictly less than 1. The remainder of the proof proceeds by showing the following two lemmas:
\begin{lemma}[Real part non-positive]\label{lem:negative} Let $\lambda$ be an eigenvalue of matrix $J_{\text{small}}$. It holds that $\textrm{Re}(\lambda) \leq 0$.
\end{lemma}

\begin{figure*}[th!]
	\centering
	\begin{tabular}{cc}
		\includegraphics[width=0.45\linewidth]{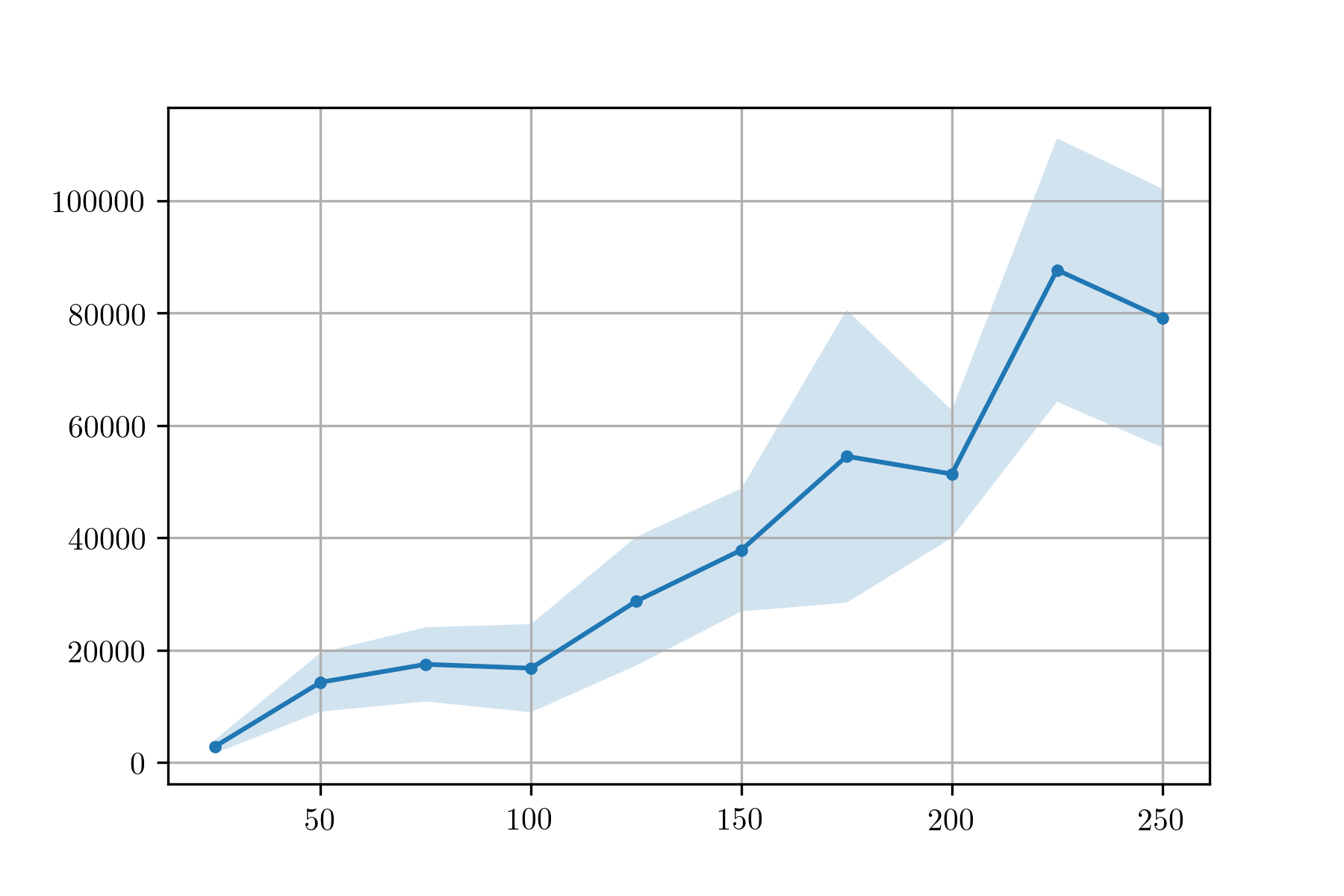}&		
		\includegraphics[width=0.45\linewidth]{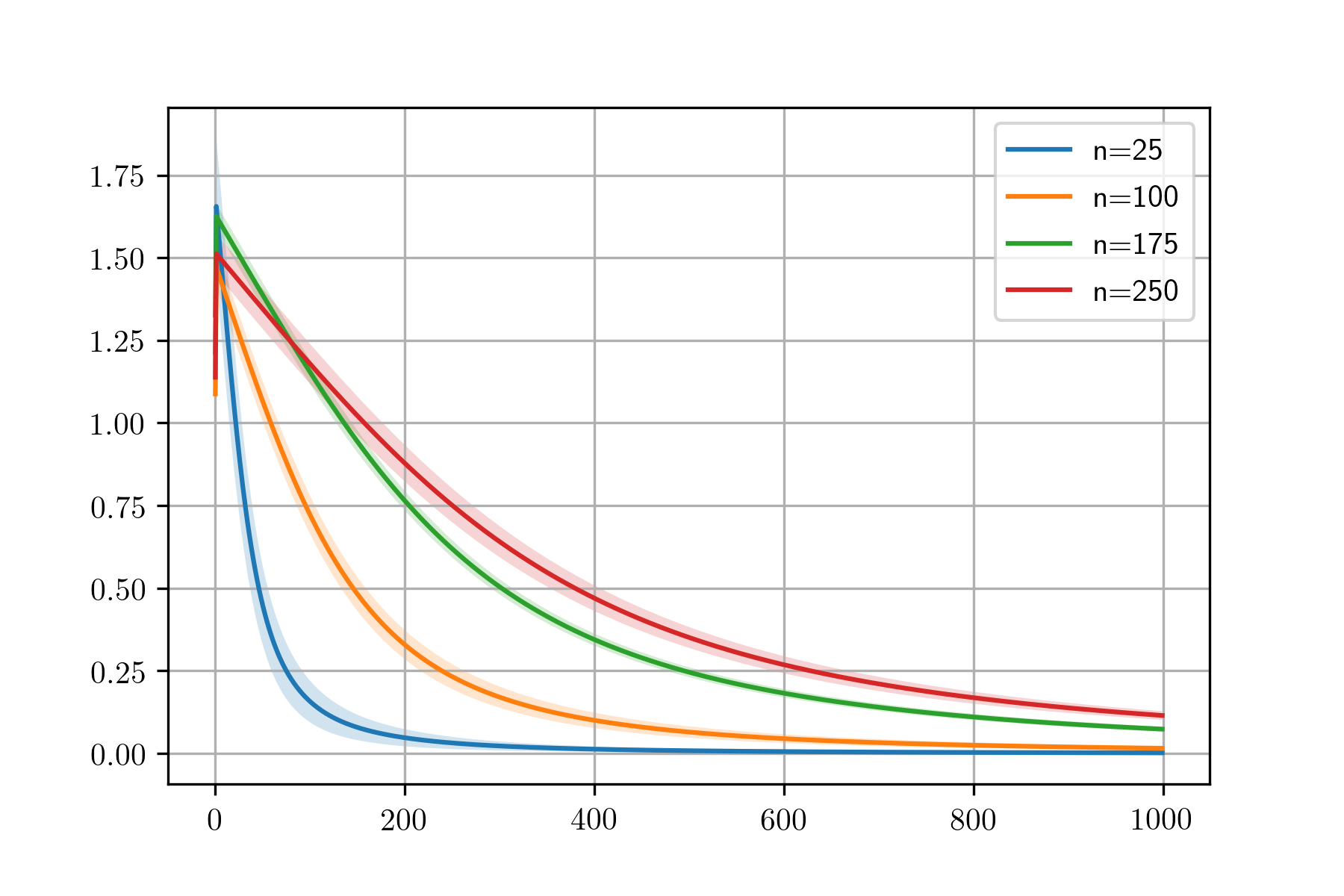}\\
		(a) \#iterations vs size of $n$ & (b) $l_1$ error vs \#iterations
	\end{tabular}	
	\caption{{\em Convergence of OMWU vs different sizes of the problem.} For Figure (a), $x$-axis is $n$ and $y$-axis is the number of iterations to reach convergence for Eqn. \eqref{eqn:linear}. In Figure (b) we choose four cases of $n$ to illustrate how $l_1$ error of the problem decreases with the number of iterations. }
	\label{fig:size}	
\end{figure*} 
\begin{proof}
Assume that $\lambda \neq 0$ (we exclude from the analysis the all-ones vector for $\vec{x}$ and for $\vec{y}$ which are the two only eigenvectors of $J_{\textrm{small}}$ with eigenvalue zero; note this is true because we assume $H$ is invertible, Remark \ref{rem:support}. All the non-zero eigenvalues of matrix $J_{\text{small}}$ coincide with the eigenvalues of the matrix 
\[
R: = \left[
\begin{array}{cc}
(D_{\vec{x}^*}-\vec{x}^{*}\vec{x}^{*\top}) & \vec{0}_{k_x \times k_y}
\\
\vec{0}_{k_y \times k_x} & (D_{\vec{y}^*}-\vec{y}^{*}\vec{y}^{*\top})
\end{array}
\right]^{1/2} \times H^{-} \times
\left[ 
\begin{array}{cc}
(D_{\vec{x}^*}-\vec{x}^{*}\vec{x}^{*\top}) & \vec{0}_{k_x \times k_y}
\\
\vec{0}_{k_y \times k_x} & (D_{\vec{y}^*}-\vec{y}^{*}\vec{y}^{*\top})
\end{array}
\right]^{1/2}.
\]
This is well-defined since \[
\left[
\begin{array}{cc}
(D_{\vec{x}^*}-\vec{x}^{*}\vec{x}^{*\top}) & \vec{0}_{k_x \times k_y}
\\
\vec{0}_{k_y \times k_x} & (D_{\vec{y}^*}-\vec{y}^{*}\vec{y}^{*\top})
\end{array}
\right]\] is positive semi-definite. 
Moreover, we use KyFan inequalities which state that the sequence (in decreasing order) of the eigenvalues of $\frac{1}{2}(W+W^{\top})$ majorizes the real part of the sequence of the eigenvalues of $W$ for any square matrix $W$ (see \cite{kyfan}, page 4). We conclude that for any eigenvalue $\lambda$ of $R$, it holds that $\textrm{Re}(\lambda)$ is at most the maximum eigenvalue of $\frac{1}{2} (R+R^{\top})$. Observe now that
\[
R+R^{\top}: = \left[
\begin{array}{cc}
(D_{\vec{x}^*}-\vec{x}^{*}\vec{x}^{*\top}) & \vec{0}_{k_x \times k_y}
\\
\vec{0}_{k_y \times k_x} & (D_{\vec{y}^*}-\vec{y}^{*}\vec{y}^{*\top})
\end{array}
\right]^{1/2} \times 
\]
\[
(H^{-}+H^{- \top}) \times
\left[
\begin{array}{cc}
(D_{\vec{x}^*}-\vec{x}^{*}\vec{x}^{*\top}) & \vec{0}_{k_x \times k_y}
\\
\vec{0}_{k_y \times k_x} & (D_{\vec{y}^*}-\vec{y}^{*}\vec{y}^{*\top})
\end{array}
\right]^{1/2}.
\]
Since 
\[
H^{-}+H^{- \top} =
\left[
\begin{array}{cc}
-\nabla_{\vec{x}\vec{x}}^2f & 0\\
0 & \nabla_{\vec{y}\vec{y}}^2f
\end{array}
\right]
\]
by the convex-concave assumption on $f$ it follows that the matrix above is negative semi-definite (see Remark \ref{rem:support}) and so is $R+R^{\top}$. We conclude that the maximum eigenvalue of $R+R^{\top}$ is non-positive. Therefore any eigenvalue of $R$ has real part non-positive and the same is true for $J_{\textrm{small}}$.
\end{proof}

\begin{lemma}\label{lem:jsmall}
	If $\epsilon$ is a non-zero eigenvalue of $J_{\text{small}}$ then, $\textrm{Re}(\epsilon) \leq 0$ and $\abs{\epsilon} \downarrow 0$ as the stepsize $\eta \to 0$.
\end{lemma}

\begin{figure}[h]
\centering
\includegraphics[width=0.45\textwidth]{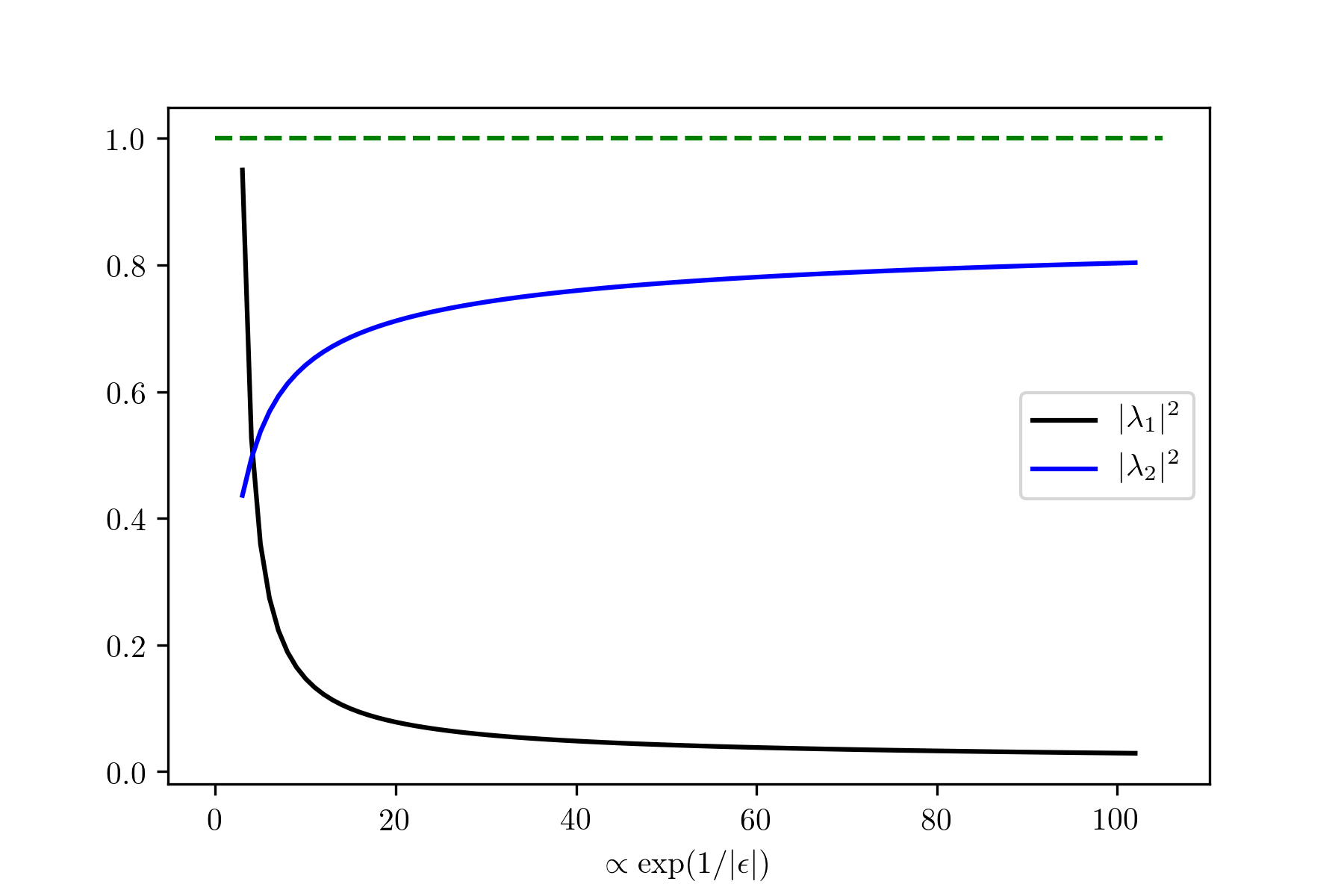}
\caption{$\lambda_1$ and $\lambda_2$ less than 1 as $\abs{\epsilon}$ is small.}
\label{fig:epsilon}
\end{figure}

We first can see that $\eta$ which is the learning rate multiplies any eigenvalue and we may assume that whilst $\eta$ is positive, it may be chosen to be sufficiently small and hence the magnitude of any eigenvalue $\abs{\epsilon} \downarrow 0$.
\begin{remark}
The equation $\epsilon=\frac{\lambda(\lambda-1)}{2\lambda-1}$ determines two complex roots for each fixed $\epsilon$, say $\lambda_1$ and $\lambda_2$. The relation between $\abs{\epsilon}$, $\abs{\lambda_1}$ and $|\lambda_2|$ is illustrated in Figure \ref{fig:epsilon}, where the $x$-axis is taken to be $\propto \exp(1/\abs{\epsilon})$. Specifically we choose $\epsilon=-1/\log(x)+1/\log(x)\sqrt{-1}$ that satisfies $|\epsilon|\downarrow 0$ as $x\rightarrow \infty$ (The $x$-axis of Figure \ref{fig:epsilon} takes $x$ from 3 to 103). 
\end{remark}

\begin{figure}[h]
\centering
\includegraphics[width=0.3\columnwidth]{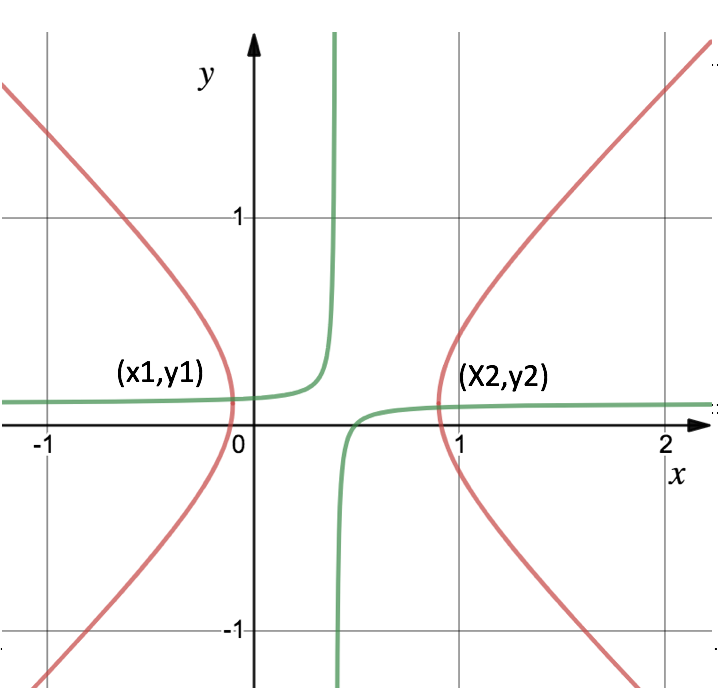}
\caption{$a=-0.1$, $b=0.1$}
\label{r}
\end{figure}

\begin{proof}
Let $\lambda=x+\sqrt{-1}y$ and $\epsilon=a+\sqrt{-1}b$. The relation $\frac{\lambda(\lambda-1)}{2\lambda-1}=\epsilon$ gives two equations based on the equality of real and imaginary parts as follows,
\begin{align}\label{eq:hp1}
x^2-x-y^2&=2ax-a-2by
\\
2xy-y&=2bx+2ay-b.
\end{align}
Notice that the above equations can be transformed to the following forms:
\begin{align}\label{eq:hp2}
(x-\frac{2a+1}{2})^2-(y-b)^2&=-a-b^2+\frac{(2a+1)^2}{4}
\\
(x-\frac{2a+1}{2})(y-b)&=ab.
\end{align}
For each $\epsilon=a+\sqrt{-1}b$, there exist two pairs of points $(x_1,y_1)$ and $(x_2,y_2)$ that are the intersections of the above two hyperbola, illustrated in Figure \ref{r}.
Recall the condition that $a<0$. As $\abs{\epsilon}\rightarrow 0$, the hyperbola can be obtained from the translation by $(\frac{2a+1}{2},b)$ of the hyperbola
\begin{align*}
x^2-y^2&=-a-b^2+\frac{(2a+1)^2}{4}
\\
xy&=ab
\end{align*}
where the translated symmetric center is close to $(\frac{1}{2},0)$ since $(a,b)$ is close to $(0,0)$. So the two intersections of the above hyperbola, $(x_1,y_1)$ and $(x_2,y_2)$, satisfy the property that $x_1^2+y_1^2$ is small and $x_2>\frac{1}{2}$ since the two intersections are on two sides of the axis $x=\frac{2a+1}{2}$, as showed in Figure \ref{intersections}.
\begin{figure}[h]
\centering
\begin{subfigure}{0.4\linewidth}
\centering
\includegraphics[width=0.6\textwidth]{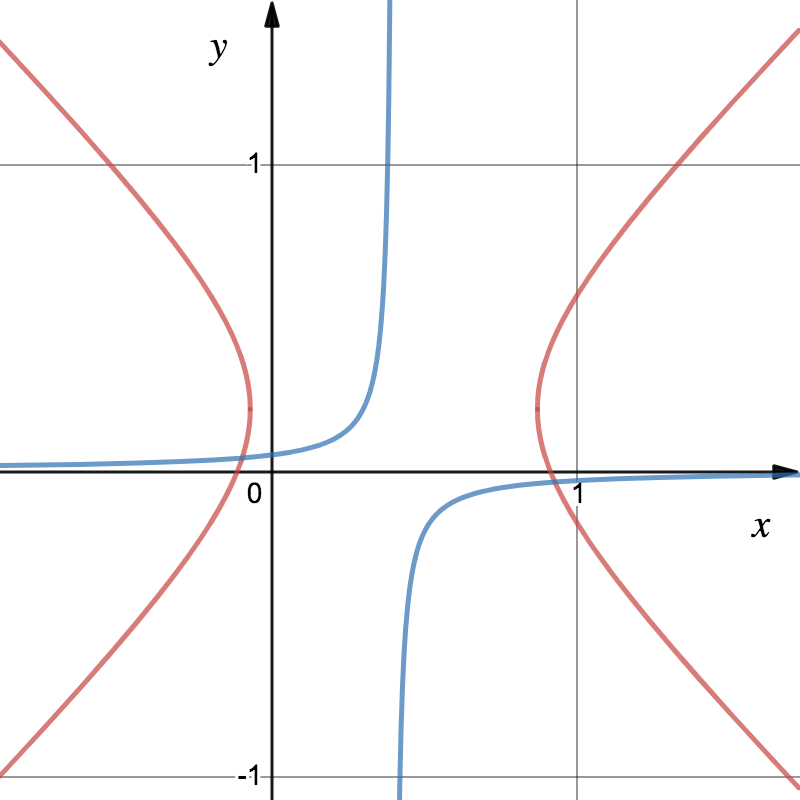}
\caption{$ab<0$}
\end{subfigure}
\ \ \ \ \ \ \ \ 
\begin{subfigure}{0.4\linewidth}
\centering
\includegraphics[width=0.6\textwidth]{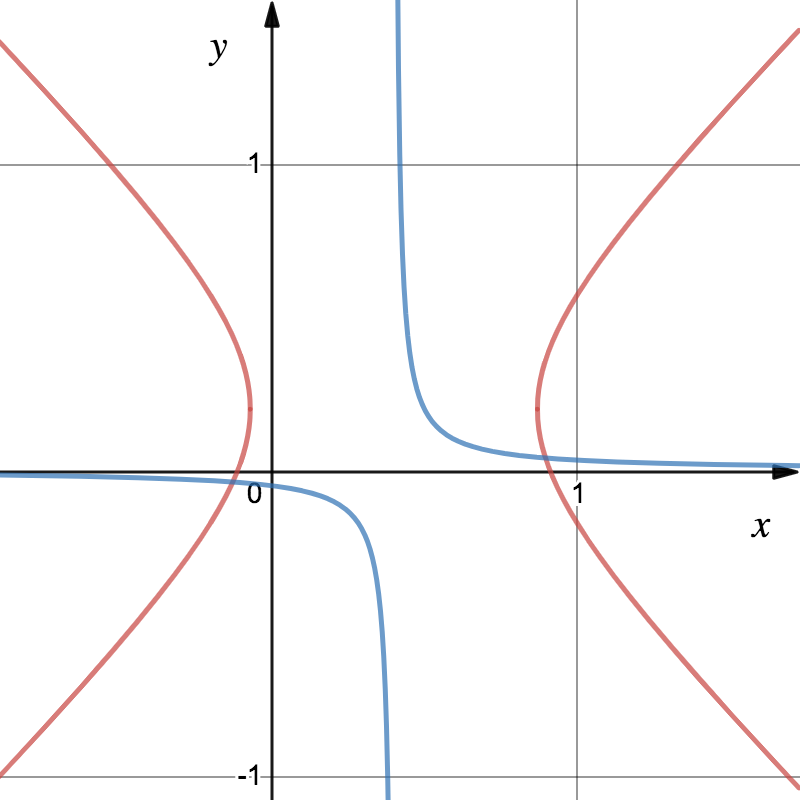}
\caption{$ab>0$}
\end{subfigure}
\caption{The intersections of the four branches of hyperbola are the two solutions of the equations (\ref{eq:hp1}) or (\ref{eq:hp2}). The intersections are on two sides of the line defined by $x=\frac{2a+1}{2}$, provided $\abs{b}$ is small and $a<0$. This occurs in the case either $ab>0$ or $ab<0$.}
\label{intersections}
\end{figure}
 On the other hand, we have
\[
\frac{\lambda(\lambda-1)}{2\lambda-1}=\frac{(x+\sqrt{-1}y)(x-1+\sqrt{-1}y)}{2x-1+\sqrt{-1}2y}=\epsilon=a+\sqrt{-1}b
\]
and then the condition $a<0$ gives the inequality
\[
\text{Re}(\epsilon)=\frac{(x^2-x+y^2)(2x-1)}{(2x-1)^2+4y^2}<0
\]
that is equivalent to 
\[
x>\frac{1}{2} \ \ \ \text{and} \ \ \ x^2-x+y^2<0
\]
where only the case $x>\frac{1}{2}$ is considered since if the intersection whose $x$-component satisfying $x<\frac{1}{2}$ has the property that $x^2+y^2$ is small and then less than 1, Figure \ref{r}. Thus to prove that $\abs{\lambda}<1$, it suffices to assume $x>\frac{1}{2}$. It is obvious that $x^2-x+y^2=(x-\frac{1}{2})^2+y^2-\frac{1}{4}<0$ implies that $x^2+y^2<1$. The proof completes. 
\end{proof}

\color{black}

\begin{figure*}[h]
	\centering
		\begin{tabular}{cc}
	\includegraphics[width=0.45\linewidth]{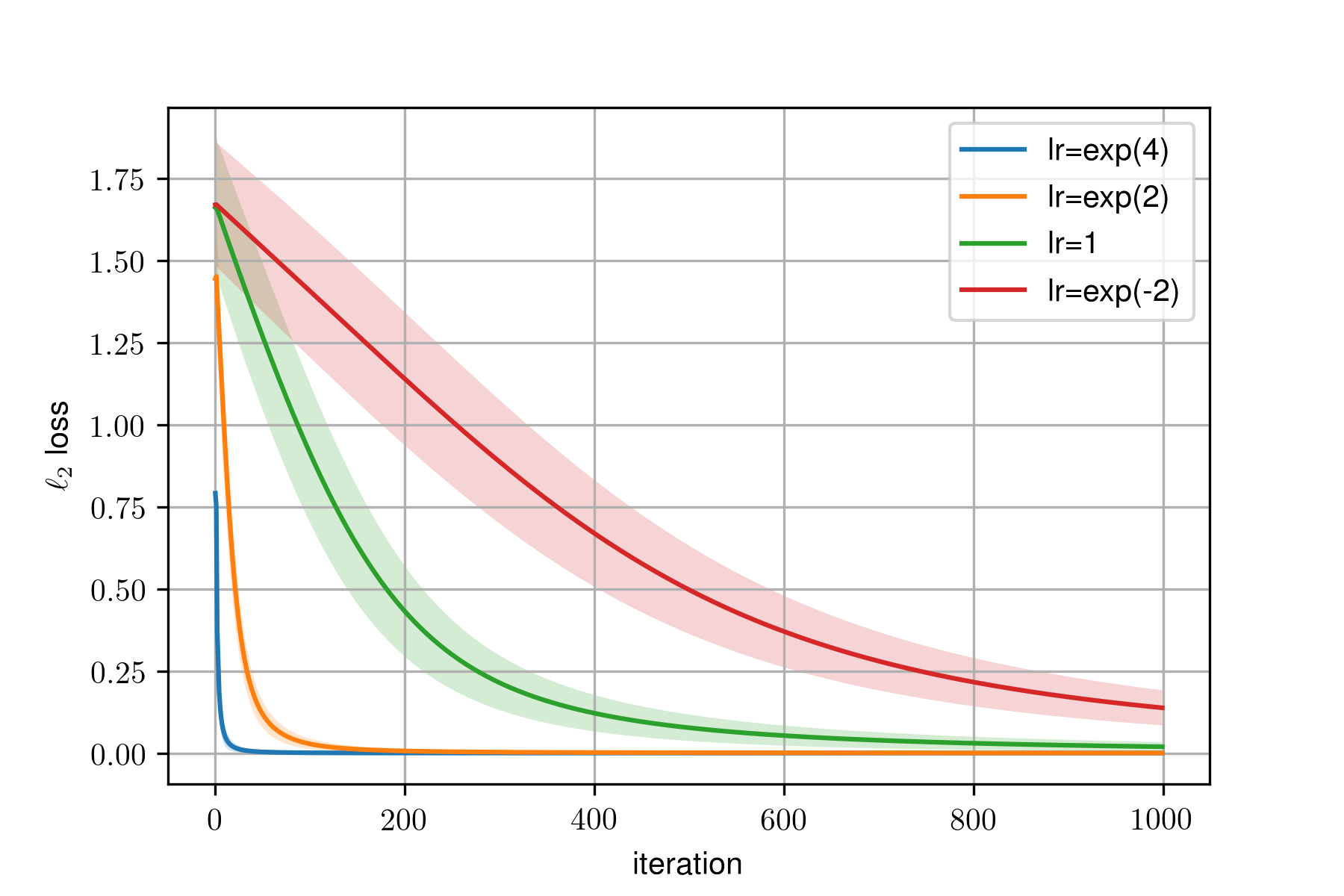}&		
			\includegraphics[width=0.45\linewidth]{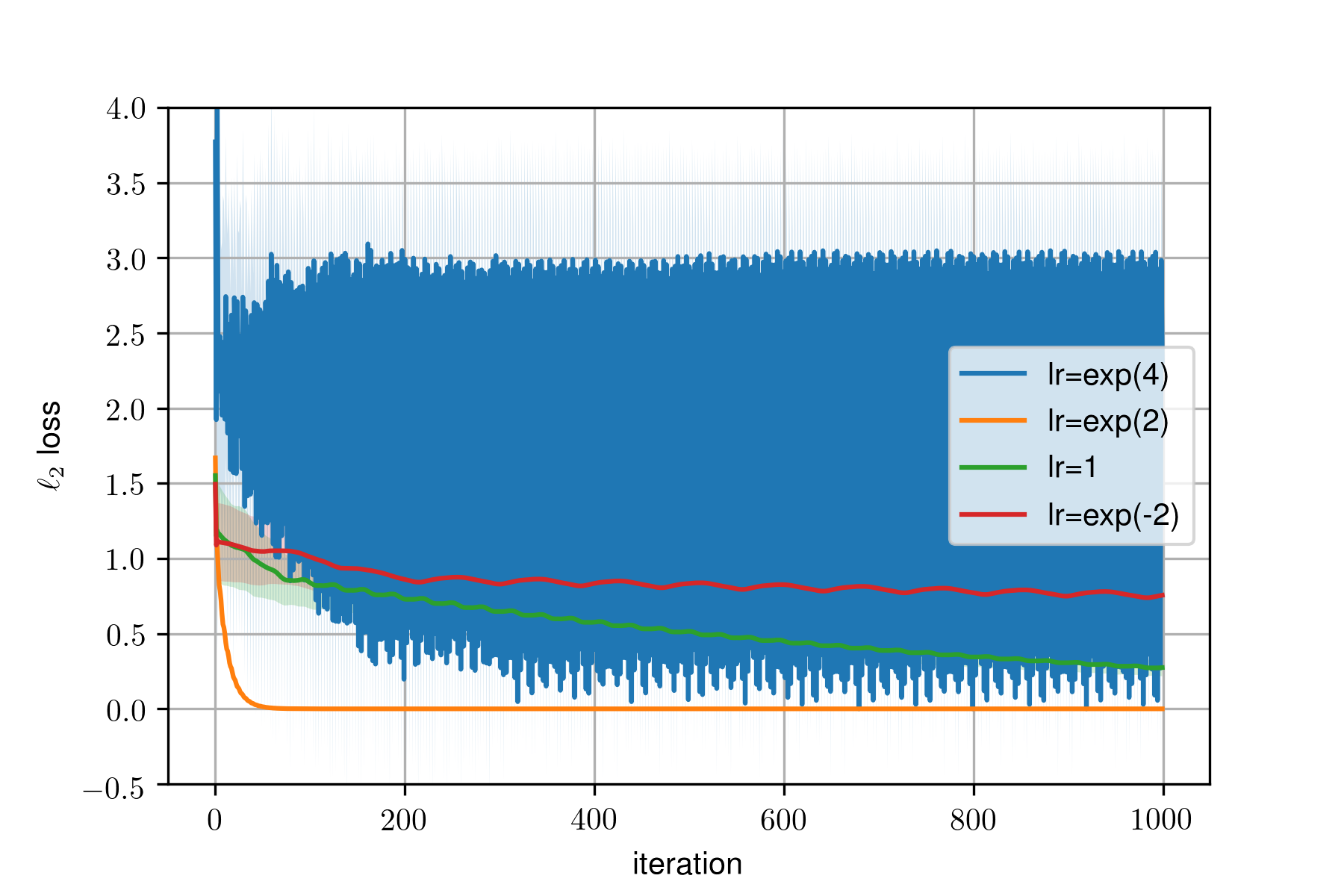} \\
			(a) OMWU & (b) OGDA\\
	\includegraphics[width=0.45\linewidth]{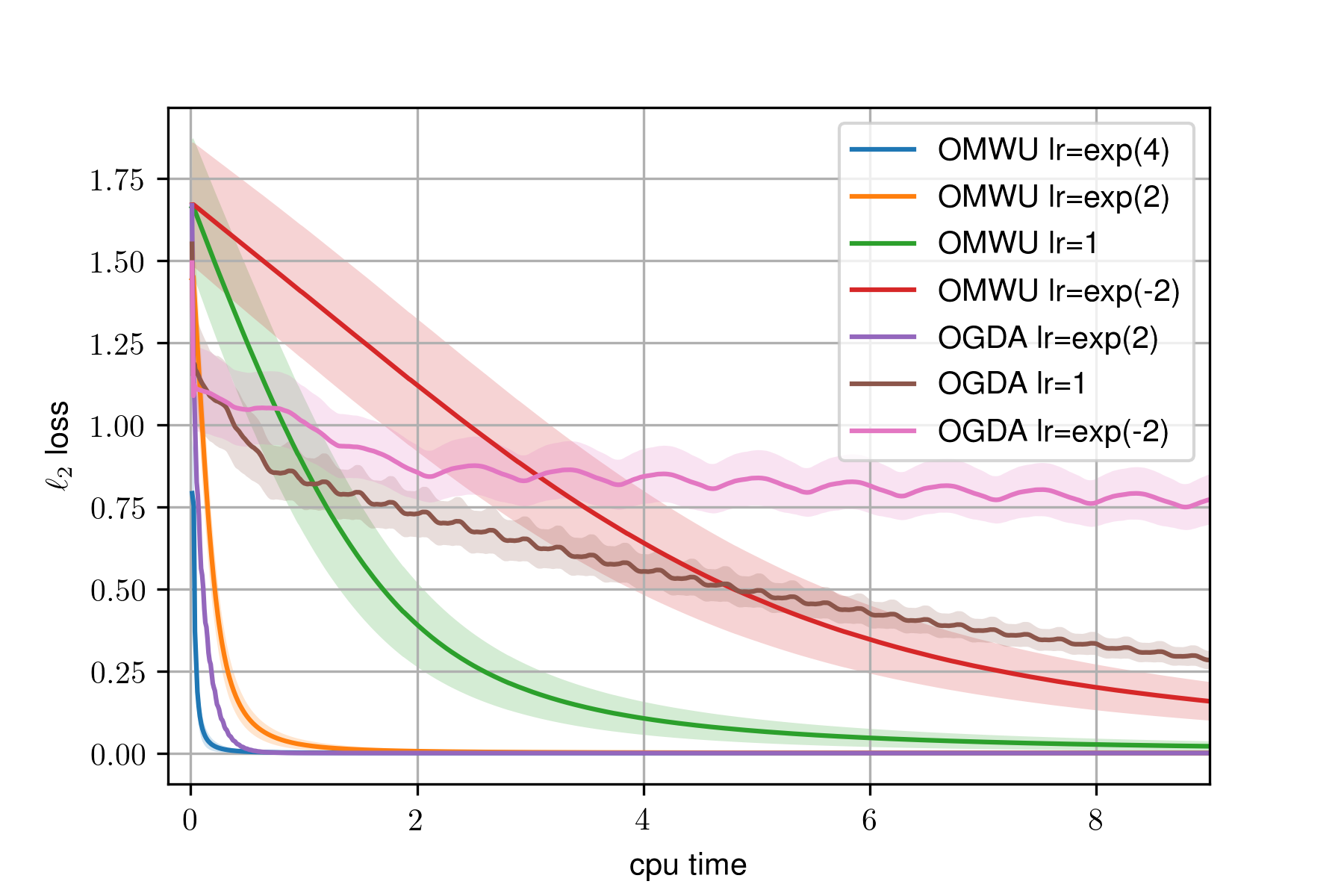}&		
			\includegraphics[width=0.45\linewidth]{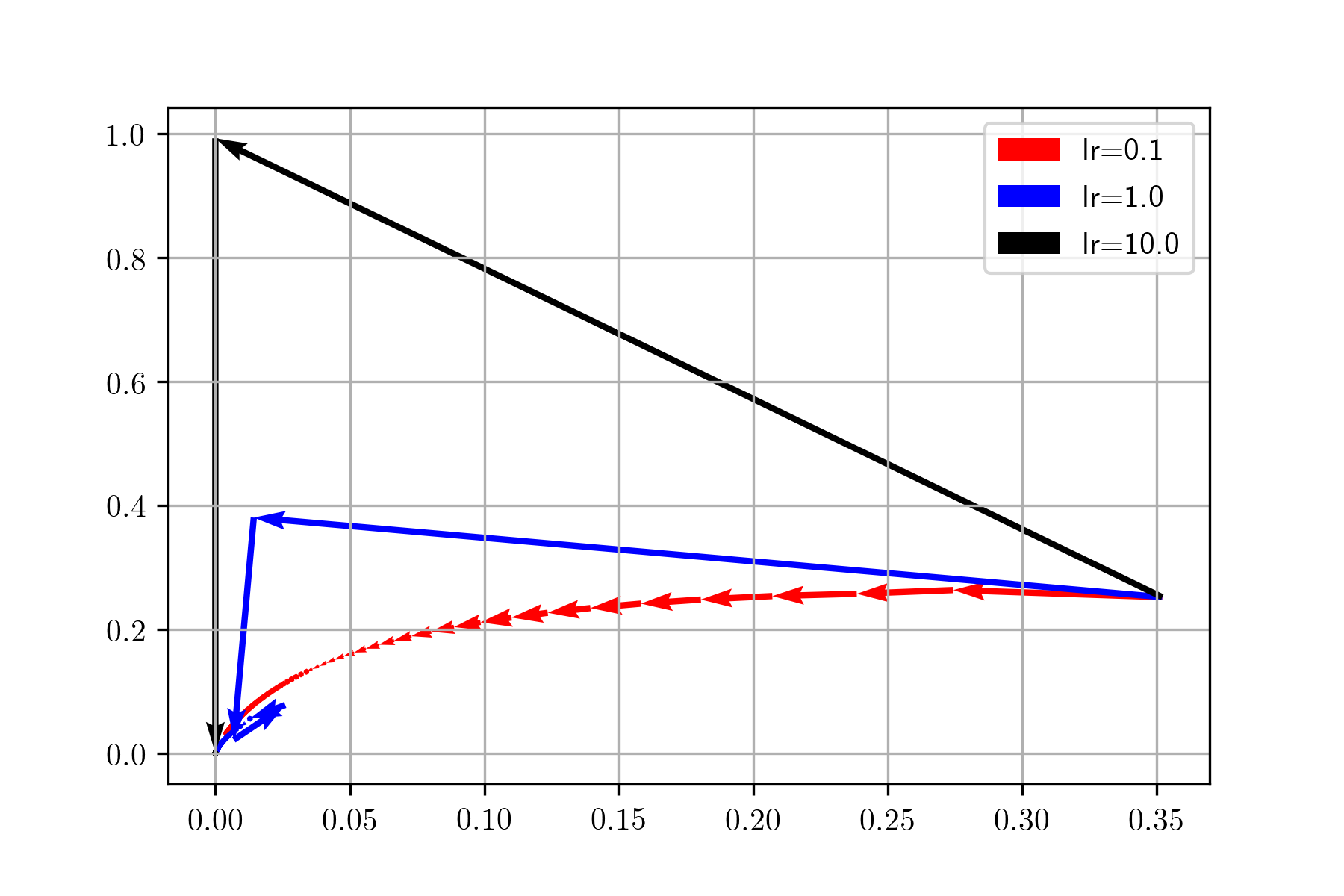}\\
			(c) Convergence time comparisons & (d) OMWU trajectories with different learning rate
		\end{tabular}	
	\caption{{\em Time comparisons of OMWU and projected OGDA vs different choices of learning rate.} For Figure (a)(b)(c), $x$-axis is iterations and $y$-axis is the $l_1$ error to the stationary point for Eqn. \eqref{eqn:linear} with $n=100$. We observe that OMWU (as in (a)) always converges while projected OGDA (as in (b)) will diverge for large learning rate. In figure (c) we remove the divergent case and compare the efficiency of the two algorithm measured in CPU time. In Figure (d) we visually present the trajectories for the min-max game of $\min_{\vec{x}\in \Delta_2}\max_{\vec{y}\in\Delta_2}\{x_1^2-y_1^2+2x_1y_1\}$ with learning rate $0.1,1.0$ and $10$. Here $x$-axis is the value of $x_1$ and $y$-axis is the value of $y_1$ respectively. The equilibrium point the algorithm converges to is $\vec{x}=[0,1],\vec{y}=[0,1]$.}
	\label{fig:lr}	
\end{figure*}

\begin{figure*}[htb!]
	\centering
		\begin{tabular}{cc}
			\includegraphics[width=0.45\linewidth]{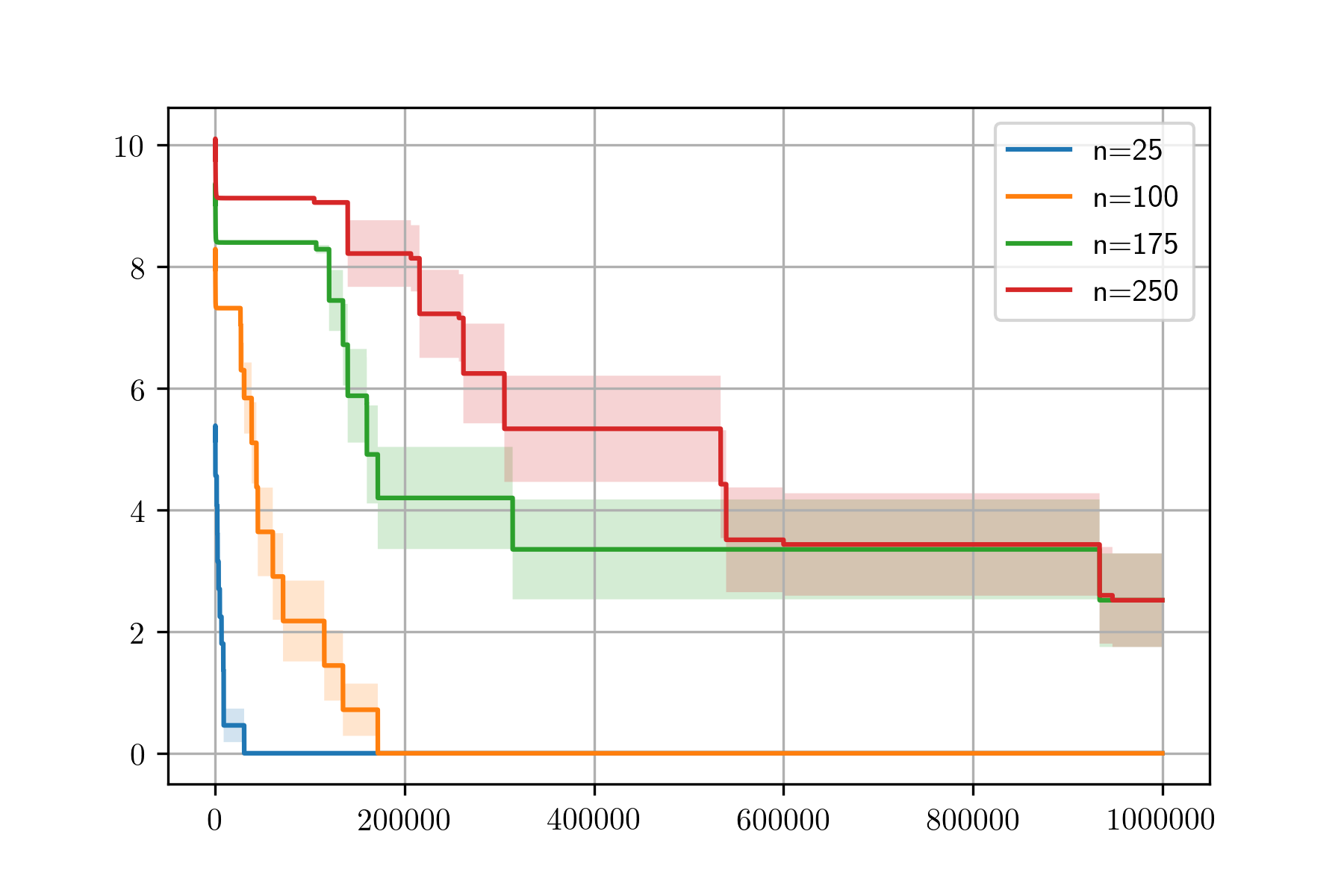}&		
			\includegraphics[width=0.45\linewidth]{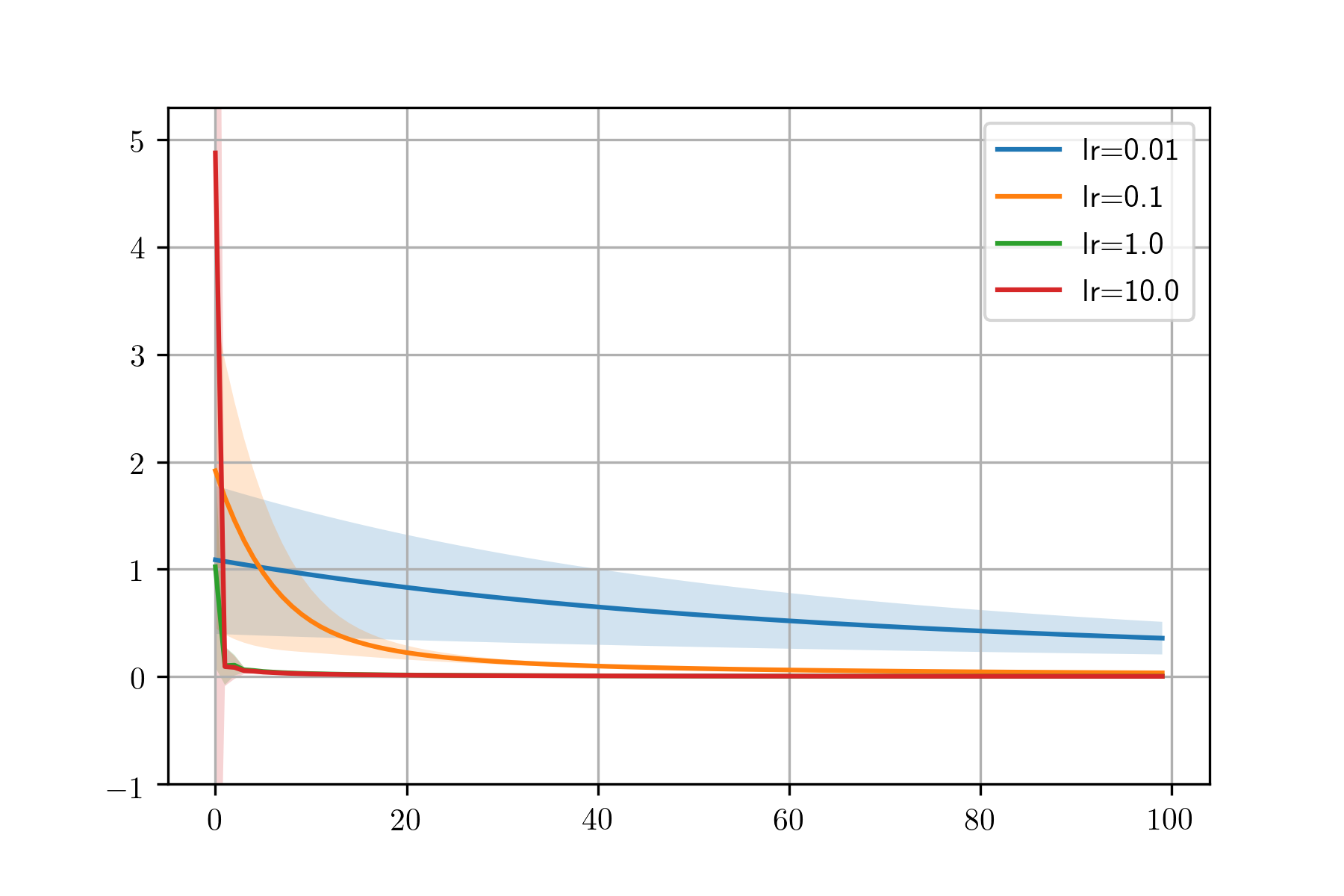}\\
			(a) KL divergence vs \#iterations with different $n$ & (b) KL divergence vs \#iterations with different $\eta$
		\end{tabular}	
	\caption{{\em KL divergence decreases with \#iterations under different settings.} For both images, $x$-axis is the number of iterations, and $y$-axis is KL divergence. Figure (a) is OMWU on bilinear function Eqn.\eqref{eqn:linear} with $n=\{25,100,175,250\}$. Figure (b) is OMWU on the quadratic function $f(\vec{x},\vec{y})=x_1^2-y_1^2+2x_1y_1$ with different learning rate $\eta$ in $\{0.01,0.1,1.0,10.0\}$. Shaded area indicates standard deviation from 10 runs with random initializations. OMWU with smaller learning rate tends to have higher variance. }	
	\label{fig:KL}
\end{figure*}

\section{Experiments}
In this section, we conduct empirical studies to verify the theoretical results of our paper. We primarily target to understand two factors that influence the convergence speed of OMWU: the problem size and the learning rate. We also compare our algorithm with Optimistic Gradient Descent Ascent (OGDA) with projection, and demonstrate our superiority against it.

We start with a simple bilinear min-max game:
\begin{eqnarray}
\label{eqn:linear}
\min_{\vec{x}\in \Delta_n}\max_{\vec{y}\in\Delta_n} \vec{x}^\top A \vec{y}.
\end{eqnarray}
We first vary the value of $n$ to study how the learning speed scales with the size of the problem. The learning rate is fixed at $1.0$, and we run OMWU with $n\in \{25,50,75,\cdots,250\}$ and matrix $A\in \mathbb{R}^{n\times n}$ is generated with i.i.d random Gaussian entries. We output the number of iterations for OMWU to reach convergence, i.e., with $l_1$ error to the optimal solution to be less or equal to $10^{-5}$. The results are averaged from 10 runs with different randomly initializations. As reported in Figure \ref{fig:size}, generally a larger problem size requires more iterations to reach convergence. We also provide four specific cases of $n$ to show the convergence in $l_1$ distance in Figure \ref{fig:size}(b). The shaded area demonstrates the standard deviation from the 50 runs.  

To understand how learning rate affects the speed of convergence, we conduct similar experiments on Eqn. \eqref{eqn:linear} and plot the $l_1$ error with different step sizes in Figure \ref{fig:lr}(a)-(c). For this experiment the matrix size is fixed as $n=100$. We also include a comparison with the Optimistic Gradient Descent Ascent\cite{DP18}. Notice the original proposal was for unconstrained problems and we use projection in each step in order to constrain the iterates to stay inside the simplex. For the setting we considered, we observe a larger learning rate effectively speeds up our learning process, and our algorithm is relatively more stable to the choice of step-size. In comparison, OGDA is quite sensitive to the choice of step-size. As shown in Figure \ref{fig:lr}(b), a larger step-size makes the algorithm diverge, while a smaller step-size will make very little progress. Furthermore, we also choose to perform our algorithm over a convex-concave but not bilinear function $f(\vec{x},\vec{y})=x_1^2-y_1^2+2x_1y_1$, where $\vec{x},\vec{y}\in \Delta_2$ and $x_1$ and $y_1$ are the first coefficients of $\vec{x}$ and $\vec{y}$. With this low dimensional function, we could visually show the convergence procedure as in Figure \ref{fig:lr}(b), where each arrow indicates an OMWU step. This figure demonstrates that at least in this case, a larger step size usually makes sure a bigger progress towards the optimal solution. 

Finally we show how the KL divergence $D_{KL}((\vec{x}^*,\vec{y}^*)\parallel(\vec{x}^{t},\vec{y}^t))$ decreases under different circumstances. Figure \ref{fig:KL} again considers the bilinear problem (Eqn.\eqref{eqn:linear}) with multiple dimensions $n$ and a simple convex-concave function $f(\vec{x},\vec{y})=x_1^2-y_1^2+2x_1y_1$ with different learning rate. We note that in all circumstances we consider, we observe that OMWU is very stable, and achieves global convergence invariant to the problem size, random initialization, and learning rate.

\section{Conclusion}
In this paper we analyze the last iterate behavior of a no-regret learning algorithm called Optimistic Multiplicative Weights Update for convex-concave landscapes. We prove that OMWU exhibits last iterate convergence in a neighborhood of the fixed point of OMWU algorithm, generalizing previous results that showed last iterate convergence for bilinear functions. The experiments explores how the problem size and the choice of learning rate affect the performance of our algorithm. We find that OMWU achieves global convergence and less sensitive to the choice of hyperparameter, compared to projected optimistic gradient descent ascent.

\bibliography{sigproc3}
\bibliographystyle{plain}
\newpage
\appendix
\section{Equations of the Jacobian of OMWU}
\begin{align}
\frac{\partial g_{1,i}}{\partial x_i}&=\frac{e^{-2\eta\frac{\partial f}{\partial x_i}+\eta\frac{\partial f}{\partial z_i}}}{S_x}+x_i\frac{1}{S_x^2}\left(e^{-2\eta\frac{\partial f}{\partial x_i}+\eta\frac{\partial f}{\partial z_i}}(-2\eta\frac{\partial^2f}{\partial x_i^2})S_x-e^{-2\eta\frac{\partial f}{\partial x_i}+\eta\frac{\partial f}{\partial z_i}}\frac{\partial S_x}{\partial x_i}\right)
\\
&\text{where}\ \ \frac{\partial S_x}{\partial x_i}=e^{-2\eta\frac{\partial f}{\partial x_i}+\eta\frac{\partial f}{\partial z_i}}-2\eta\sum_k x_ke^{-2\eta\frac{\partial f}{\partial x_k}+\eta\frac{\partial f}{\partial z_k}}\frac{\partial^2f}{\partial x_i^2}
\\
\frac{\partial g_{1,i}}{\partial x_j}&=x_i\frac{1}{S_x^2}\left(e^{-2\eta\frac{\partial f}{\partial x_i}+\eta\frac{\partial f}{\partial z_i}}(-2\eta\frac{\partial^2f}{\partial x_i\partial x_j})S_x-e^{-2\eta\frac{\partial f}{\partial x_i}+\eta\frac{\partial f}{\partial z_i}}\frac{\partial S_x}{\partial x_j}\right)
\\
&\text{where}\ \ \frac{\partial S_x}{\partial x_j}=e^{-2\eta\frac{\partial f}{\partial x_j}+\eta\frac{\partial f}{\partial z_j}}-2\eta\sum_k x_ke^{-2\eta\frac{\partial f}{\partial x_k}+\eta\frac{\partial f}{\partial z_k}}\frac{\partial^2f}{\partial x_j\partial x_k}
\\
\frac{\partial g_{1,i}}{\partial y_j}&=x_i\frac{1}{S_x^2}\left(e^{-2\eta\frac{\partial f}{\partial x_i}+\eta\frac{\partial f}{\partial z_i}}(-2\eta\frac{\partial^2f}{\partial x_i\partial y_j})S_x-e^{-2\eta\frac{\partial f}{\partial x_i}+\eta\frac{\partial f}{\partial z_i}}\frac{\partial S_x}{\partial y_j}\right)
\\
&\text{where}\ \ \frac{\partial S_x}{\partial y_j}=\sum_k x_ke^{-2\eta\frac{\partial f}{\partial x_i}+\eta\frac{\partial f}{\partial z_i}}(-2\eta\frac{\partial^2f}{\partial x_k\partial y_j})
\\
\frac{\partial g_{1,i}}{\partial z_j}&=x_i\frac{1}{S_x^2}\left(e^{-2\eta\frac{\partial f}{\partial x_i}+\eta\frac{\partial f}{\partial z_i}}(\eta\frac{\partial^2f}{\partial z_j\partial x_i})S_x-e^{-2\eta\frac{\partial f}{\partial x_i}+\eta\frac{\partial f}{\partial z_i}}\frac{\partial S_x}{\partial z_j}\right)
\\
&\text{where}\ \ \frac{\partial S_x}{\partial z_j}=\eta\sum_k x_ke^{-2\eta\frac{\partial f}{\partial x_k}+\eta\frac{\partial f}{\partial z_k}}\frac{\partial^2f}{\partial z_k\partial z_j}
\\
\frac{\partial g_{1,i}}{\partial w_j}&=x_i\frac{1}{S_x^2}\left(e^{-2\eta\frac{\partial f}{\partial x_i}+\eta\frac{\partial f}{\partial z_i}}\eta\frac{\partial^2f}{\partial z_i\partial w_j}S_x-e^{-2\eta\frac{\partial f}{\partial x_i}+\eta\frac{\partial f}{\partial z_i}}\frac{\partial S_x}{\partial w_j}\right)
\\
&\text{where}\ \ \frac{\partial S_x}{\partial w_j}=\sum_k x_ke^{-2\eta\frac{\partial f}{\partial x_k}+\eta\frac{\partial f}{\partial z_k}}\eta\frac{\partial f}{\partial z_k\partial w_j}
\\
\frac{\partial g_{2,i}}{\partial x_j}&=y_i\frac{1}{S_y^2}\left(e^{2\eta\frac{\partial f}{\partial y_i}-\eta\frac{\partial f}{\partial w_i}}(2\eta\frac{\partial^2f}{\partial x_j\partial y_i})S_y-e^{2\eta\frac{\partial f}{\partial y_i}-\eta\frac{\partial f}{\partial w_i}}\frac{\partial S_y}{\partial x_j}\right)
\\
&\text{where}\ \ \frac{\partial S_y}{\partial x_j}=\sum_ky_ke^{2\eta\frac{\partial f}{\partial y_i}-\eta\frac{\partial f}{\partial w_i}}2\eta\frac{\partial^2f}{\partial x_j\partial y_k}
\\
\frac{\partial g_{2,i}}{\partial y_i}&=\frac{e^{2\eta\frac{\partial f}{\partial y_i}-\eta\frac{\partial f}{\partial w_i}}}{S_y}+y_i\frac{1}{S_y^2}\left(e^{2\eta\frac{\partial f}{\partial y_i}-\eta\frac{\partial f}{\partial w_i}}2\eta\frac{\partial^2f}{\partial y_i^2}S_y-e^{2\eta\frac{\partial f}{\partial y_i}-\eta\frac{\partial f}{\partial w_i}}\frac{\partial S_y}{\partial y_i}\right)
\\
&\text{where}\ \ \frac{\partial S_y}{\partial y_i}=e^{2\eta\frac{\partial f}{\partial y_i}-\eta\frac{\partial f}{\partial w_i}}+2\eta\sum_ky_ke^{2\eta\frac{\partial f}{\partial y_i}-\eta\frac{\partial f}{\partial w_i}}\frac{\partial^2f}{\partial y_i\partial y_k}
\\
\frac{\partial g_{2,i}}{\partial z_j}&=y_i\frac{1}{S_y^2}\left(e^{2\eta\frac{\partial f}{\partial y_i}-\eta\frac{\partial f}{\partial w_i}}(-\eta\frac{\partial^2f}{\partial w_i\partial z_j})S_y-e^{2\eta\frac{\partial f}{\partial y_i}-\eta\frac{\partial f}{\partial w_i}}\frac{\partial S_y}{\partial z_j}\right)
\\
&\text{where}\ \ \frac{\partial S_y}{\partial z_j}=\sum_ky_ke^{2\eta\frac{\partial f}{\partial y_i}-\eta\frac{\partial f}{\partial w_i}}(-\eta\frac{\partial^2f}{\partial w_k\partial z_j})
\\
\frac{\partial g_{2,i}}{\partial w_j}&=y_i\frac{1}{S_y^2}\left(e^{2\eta\frac{\partial f}{\partial y_i}-\eta\frac{\partial f}{\partial w_i}}(-\eta\frac{\partial^2f}{\partial w_i\partial w_j})-e^{2\eta\frac{\partial f}{\partial y_i}-\eta\frac{\partial f}{\partial w_i}}\frac{\partial S_y}{\partial w_j}\right)
\\
&\text{where}\ \ \frac{\partial S_y}{\partial w_j}=\sum_ky_ke^{2\eta\frac{\partial f}{\partial y_i}-\eta\frac{\partial f}{\partial w_i}}(-\eta\frac{\partial^2f}{\partial w_k\partial w_j})
\end{align}

\newpage
\section{Equations of the Jacobian of OMWU at the fixed point $(\vec{x}^*,\vec{y}^*,\vec{z}^*,\vec{w}^*)$}
In this section, we compute the equations of the Jacobian at the fixed point $(\vec{x}^*,\vec{y}^*,\vec{z}^*,\vec{w}^*)$. The fact that $(\vec{x}^*,\vec{y}^*)=(\vec{z}^*,\vec{w}^*)$ and $(\vec{z},\vec{w})$ takes the position of $(\vec{x},\vec{y})$ in computing partial derivatives gives the following equations.
\begin{align}
\frac{\partial g_{1,i}}{\partial x_i}&=1-x_i^*-2\eta x_i^*(\frac{\partial^2f}{\partial x_i^*}-\sum_kx_k^*\frac{\partial^2f}{\partial x_i\partial x_k}), i\in[n],
\\
\frac{\partial g_{1,i}}{\partial x_j}&=-x_i^*-2\eta x_i^*(\frac{\partial^2f}{\partial x_i\partial x_j}-\sum_k x_k^*\frac{\partial^2f}{\partial x_j\partial x_k}),j\in[n],j\ne i
\\
\frac{\partial g_{1,i}}{\partial y_j}&=-2\eta x_i^*(\frac{\partial^2f}{\partial x_i\partial y_j}-\sum_k x_k^*\frac{\partial^2f}{\partial x_k\partial y_j}),j\in[m]
\\
\frac{\partial g_1,i}{\partial z_j}&=\eta x_i^*(\frac{\partial^2f}{\partial x_i\partial x_j}-\sum_k x_k^*\frac{\partial^2f}{\partial x_k\partial x_j}),j\in[n]
\\
\frac{\partial g_{1,i}}{\partial w_j}&=\eta x_i^*(\frac{\partial^2f}{\partial x_i\partial y_j}-\sum_k x_k^*\frac{\partial^2f}{\partial x_k\partial y_j}),j\in[m]
\\
\frac{\partial g_{2,i}}{\partial x_j}&=2\eta y_i^*(\frac{\partial^2f}{\partial x_j\partial y_i}-\sum_k y_k^*\frac{\partial^2f}{\partial x_j\partial y_k}),j\in[n]
\\
\frac{\partial g_{2,i}}{\partial y_i}&=1-y_i^*+2\eta(\frac{\partial^2 f}{\partial y_i^2}-\sum_ky_k^*\frac{\partial^2f}{\partial y_i\partial y_k}),i\in[m]
\\
\frac{\partial g_{2,i}}{\partial y_j}&=-y_i^*+2\eta(\frac{\partial^2f}{\partial y_i\partial y_j}-\sum_k y_k^*\frac{\partial^2f}{\partial y_j\partial y_k}),j\in[m]
\\
\frac{\partial g_{2,i}}{\partial z_j}&=\eta y_i^*(-\frac{\partial^2f}{\partial x_j\partial y_i}+\sum_k y_k^*\frac{\partial^2f}{\partial x_j\partial y_k}),j\in[n]
\\
\frac{\partial g_{2,i}}{\partial w_j}&=\eta y_i^*(-\frac{\partial^2f}{\partial y_i\partial y_j}+\sum_k y_k^*\frac{\partial^2f}{\partial y_k\partial y_j}),j\in[m]
\\
\frac{\partial g_{3,i}}{\partial x_i}&=1 \ \ \text{for all}\ \ i\in[n] \ \ \text{and zero}\text{for all the other partial derivatives of}\ \  g_{3,i}
\\
\frac{\partial g_{4,i}}{\partial y_i}&=1 \ \text{for all}\ \ i\in[m] \ \ \text{and zero
for all the other partial derivatives of}\ \ \ g_{4,i}.
\end{align}

\section{Jacobian matrix at $(\vec{x}^*,\vec{y}^*,\vec{z}^*,\vec{w}^*)$}
This section serves for the "Spectral Analysis" of Section 3. The Jacobian matrix of $g$ at the fixed point is obtained based on the calculations above. We refer the main article for the subscript indicating the size of each block matrix.
\scriptsize
\[
J=
\left[
\begin{array}{cccc}
\vec{I}-D_{\vec{x}^*}\vec{1}\vec{1}^{\top}-2\eta D_{\vec{x}^*}(\vec{I}-\vec{1}\vec{x}^{*\top})\nabla_{\vec{x}\vec{x}}^2f & -2\eta D_{\vec{x}^*}(\vec{I}-\vec{1}\vec{x}^{*\top})\nabla_{\vec{x}\vec{y}}^2f & \eta D_{\vec{x}^*}(\vec{I}-\vec{1}\vec{x}^{*\top})\nabla_{\vec{x}\vec{x}}^2f & \eta D_{\vec{x}^{*}}(\vec{I}-\vec{1}\vec{x}^{*\top})\nabla_{\vec{x}\vec{y}}^2f
\\
2\eta D_{\vec{y}^{*\top}}(\vec{I}-\vec{1}\vec{y}^{*})\nabla_{\vec{y}\vec{x}}^2f & \vec{I}-D_{\vec{y}^*}\vec{1}\vec{1}^{\top}+2\eta D_{\vec{y}^*}(\vec{I}-\vec{1}\vec{y}^{*\top})\nabla_{\vec{y}\vec{y}}^2f & -\eta D_{\vec{y}^*}(\vec{I}-\vec{1}\vec{y}^{*\top})\nabla_{\vec{y}\vec{x}}^2f & -\eta D_{\vec{y}^*}(\vec{I}-\vec{1}\vec{y}^{*\top})\nabla_{\vec{y}\vec{y}}^2f
\\
\vec{I} & \vec{0} & \vec{0} & \vec{0}
\\
\vec{0} & \vec{I} & \vec{0} & \vec{0}
\end{array}
\right]
\]

\normalsize
By acting on the tangent space of each simplex, we observe that $D_{\vec{x}^*}\vec{1}\vec{1}^{\top}\vec{v}=0$ for $\sum_k v_k=0$, so each eigenvalue of matrix $J$ is an eigenvalue of the following matrix  
\scriptsize
\[
J_{\text{new}}=
\left[
\begin{array}{cccc}
\vec{I}-2\eta D_{\vec{x}^*}(\vec{I}-\vec{1}\vec{x}^{*\top})\nabla_{\vec{x}\vec{x}}^2f & -2\eta D_{\vec{x}^*}(\vec{I}-\vec{1}\vec{x}^{*\top})\nabla_{\vec{x}\vec{y}}^2f & \eta D_{\vec{x}^*}(\vec{I}-\vec{1}\vec{x}^{*\top})\nabla_{\vec{x}\vec{x}}^2f & \eta D_{\vec{x}^{*}}(\vec{I}-\vec{1}\vec{x}^{*\top})\nabla_{\vec{x}\vec{y}}^2f
\\
2\eta D_{\vec{y}^{*}}(\vec{I}-\vec{1}\vec{y}^{*\top})\nabla_{\vec{y}\vec{x}}^2f & \vec{I}+2\eta D_{\vec{y}^*}(\vec{I}-\vec{1}\vec{y}^{*\top})\nabla_{\vec{y}\vec{y}}^2f & -\eta D_{\vec{y}^*}(\vec{I}-\vec{1}\vec{y}^{*\top})\nabla_{\vec{y}\vec{x}}^2f & -\eta D_{\vec{y}^*}(\vec{I}-\vec{1}\vec{y}^{*\top})\nabla_{\vec{y}\vec{y}}^2f
\\
\vec{I} & \vec{0} & \vec{0} & \vec{0}
\\
\vec{0} & \vec{I} & \vec{0} & \vec{0}
\end{array}
\right]
\]
\normalsize
The characteristic polynomial of $J_{\text{new}}$ is $\det (J_{new}-\lambda I)$ that can be computed as the determinant of the following matrix:
\begin{align}
&\left[
\begin{array}{cccc}
(1-\lambda)\vec{I}+(\frac{1}{\lambda}-2)\eta D_{\vec{x}^*}(\vec{I}-\vec{1}\vec{x}^{*\top})\nabla_{\vec{x}\vec{x}}^2f & (\frac{1}{\lambda}-2)\eta D_{\vec{x}^*}(\vec{I}-\vec{1}\vec{x}^{*\top})\nabla_{\vec{x}\vec{y}}^2f
\\
(2-\frac{1}{\lambda})\eta D_{\vec{y}^{*}}(\vec{I}-\vec{1}\vec{y}^{*\top})\nabla_{\vec{y}\vec{x}}^2f & (1-\lambda)\vec{I}+(2-\frac{1}{\lambda})\eta D_{\vec{y}^{*}}(\vec{I}-\vec{1}\vec{y}^{*\top})\nabla_{\vec{y}\vec{y}}^2f
\end{array}
\right]
\end{align}


\end{document}